\documentclass[10pt,twocolumn,letterpaper]{article}

\usepackage{cvpr}
\usepackage{times}
\usepackage{epsfig}
\usepackage{graphicx}
\usepackage{amsmath, mathtools}
\usepackage{amssymb}

\usepackage{multirow}
\usepackage{makecell}
\usepackage{color}
\usepackage[dvipsnames]{xcolor}
\usepackage[font=normal]{subfig}
\usepackage[export]{adjustbox}
\usepackage{setspace}
\usepackage{booktabs}
\usepackage{array}

\cvprfinalcopy 

\def\cvprPaperID{****} 

\ifcvprfinal\fi

\begin{document}

\title{NTIRE 2020 Challenge on NonHomogeneous Dehazing}
\author{
    Codruta O. Ancuti \and Cosmin Ancuti  \and Florin-Alexandru Vasluianu \and Radu Timofte 
    \and Jing Liu  \and Haiyan Wu \and Yuan Xie \and Yanyun Qu \and Lizhuang Ma
    \and Ziling Huang \and Qili Deng 
    \and Ju-Chin Chao  \and Tsung-Shan Yang \and Peng-Wen Chen \and Po-Min Hsu \and Tzu-Yi Liao \and Chung-En Sun \and Pei-Yuan Wu
    \and Jeonghyeok Do \and Jongmin Park \and Munchurl Kim
    \and Kareem Metwaly \and Xuelu Li \and  Tiantong Guo \and Vishal Monga 
    \and  Mingzhao Yu \and Venkateswararao Cherukuri
    \and Shiue-Yuan Chuang \and Tsung-Nan Lin \and David Lee \and Jerome Chang  \and Zhan-Han Wang
    \and Yu-Bang Chang  \and Chang-Hong Lin
    \and Yu Dong \and Hongyu Zhou 
    \and Xiangzhen Kong 
    \and Sourya Dipta Das \and Saikat Dutta 
    \and Xuan Zhao
    \and  Bing Ouyang \and Dennis Estrada
    \and Meiqi Wang \and Tianqi Su \and Siyi Chen 
    \and Bangyong Sun   \and Vincent Whannou de Dravo \and  Zhe Yu
    \and Pratik Narang \and Aryan Mehra  \and Navaneeth Raghunath \and Murari Mandal
}

\maketitle

\begin{abstract}
    This paper reviews the NTIRE 2020 Challenge on NonHomogeneous Dehazing of images (restoration of rich details in hazy image). 
    We focus on the proposed solutions and their results evaluated on \textbf{NH-Haze}, a novel dataset consisting of 55 pairs of real haze free and nonhomogeneous hazy images recorded outdoor. \textbf{NH-Haze} is the first realistic nonhomogeneous haze dataset that provides ground truth images.
    The nonhomogeneous haze has been produced using a professional haze generator that imitates the real conditions of haze scenes. 168 participants registered in the challenge and 27 teams competed in the final testing phase.  The proposed solutions gauge the state-of-the-art in image dehazing.
\end{abstract}

    \let\thefootnote\relax\footnotetext{Codruta O. Ancuti (codruta.ancuti@gmail.com, University Politehnica of Timisoara), C.  Ancuti, F.-A. Vasluianu, R. Timofte are the NTIRE 2020 challenge organizers, while the other authors participated in the challenge. 
    \\Appendix~\ref{sec:appendix} contains the authors' teams and affiliations.
    \\NTIRE 2020 webpage:\\\url{https://data.vision.ee.ethz.ch/cvl/ntire20/}}

\section{Introduction}
\label{sec:introduction}
    Haze is an atmospheric  phenomena produced by small floating particles which absorb and scatter the light from its propagation direction. As a result, haze influences the visibility of such scene as it generates loss of contrast of the distant objects, selective attenuation of the light spectrum, and additional noise.
    Restoring such images is important in several outdoor computer vision  applications such as visual surveillance and automatic driving assistance.

   Image dehazing is a challenging ill-posed problem that recently has drawn a significant attention in the computer vision community~\cite{Fattal_Dehazing,Tan_Dehazing,Kratz_and_Nishino_2009,Tarel_ICCV_2009,Dehaze_He_CVPR_2009,Ancuti_TIP_2013,Liu_dehazing_2018,Ancuti_NT_TIP_2020}. Most recently, an important number of CNN-based approaches~\cite{Dehazenet_2016,Ren_2016,Zhang_dehazing_2018,Liu_2019,Wang_2019} have been introduced in the literature proving high competitiveness compared with the non-learning techniques. The image dehazing research is aligned with the advances from related tasks such as image super-resolution~\cite{Timofte_2017_CVPR_Workshops,Timofte_2018_CVPR_Workshops,Blau_2018_ECCV_Workshops,Ignatov_2018_ECCV_Workshops,cai2019ntire,gu2019aim,lugmayr2019aim}, denoising~\cite{abdelhamed2019ntire,gu2019brief}or enhancement~\cite{ignatov2019ntire}.
   
    Despite this growing interest, the field still lacks standardized benchmarks to allow for evaluating objectively and quantitatively the performance of the existing dehazing techniques. Basically, a major issue preventing further developments is related to the impossibility to reliably assess the dehazing performance of a given algorithm, due to the absence of reference haze-free images (ground-truth). A key problem in collecting pairs of hazy and haze-free ground-truth images lies in the need to capture both images with identical scene illumination.
    
    First image dehazing benchmarks with groundtruth considered synthesized hazy images, employing the optical model and known depth to synthesize the haze effect. For instance, FRIDA~\cite{Tarel_2012}  dataset, designed for Advanced Driver Assistance Systems (ADAS), is a synthetic image database with 66 computer graphics generated roads scenes. Another representative synthetic dehazing dataset is D-HAZY~\cite{D_Hazy_2016} that contains 1400+ real images and corresponding depth maps used to synthesize hazy scenes based on Koschmieder's light propagation model~\cite{Koschmieder_1924}. 
    
    An important step forward in benchmarking the dehazing technique is represented by the  first challenge for single image dehazing that has been organized by NTIRE 2018~\cite{Ancuti_NTIRE_2018}. The NTIRE 2018 challenge was  based on two realistic dehazing datasets: I-HAZE ~\cite{Ancuti_IHAZE_2018} and O-HAZE~\cite{Ancuti_OHAZE_2018}. The I-HAZE includes 35 hazy images (with haze generated in a controlled indoor environment) and their corresponding ground truth (haze-free) images of the same scene while O-HAZE dataset includes 45 outdoor hazy images and corresponding ground truth (haze-free) images. Both datasets allow for full-reference quality assessment of the dehazing results.
    A second dehazing challenge has been organised by NTIRE 2019~\cite{Ancuti_NTIRE_2019} based on DenseHaze~\cite{Ancuti_DENSE_HAZE_2019}, a more challenging dehazing dataset that considers dense hazy scenes with corresponding ground truth images.
    
    The NTIRE 2020 challenge represents a step forward in benchmarking single image dehazing. It uses a novel dehazing dataset, \textbf{NH-Haze}~\cite{ancuti2020NH-HAZE} \footnote{\url{https://data.vision.ee.ethz.ch/cvl/ntire20//nh-haze/}}, that consists from 55 hazy images with  haze generated in  outdoor environments, and their corresponding ground truth (haze-free) images of the same scene. \textbf{NH-Haze} contains real outdoor scenes with non-homogeneous haze generated using a professional haze machine that imitates the real haze conditions. To the best of our knowledge this is the first dehazing dataset with non-homogeneous hazy scenes.  
    
    The evaluation was performed objectively by comparing the restored hazy images with the ground truth images.
    
    This challenge is one of the NTIRE 2020 associated challenges on: deblurring~\cite{nah2020ntire}, nonhomogeneous dehazing~\cite{ancuti2020ntire}, perceptual extreme super-resolution~\cite{zhang2020ntire}, video quality mapping~\cite{fuoli2020ntire}, real image denoising~\cite{abdelhamed2020ntire}, real-world super-resolution~\cite{lugmayr2020ntire}, spectral reconstruction from RGB image~\cite{arad2020ntire} and demoireing~\cite{yuan2020demoireing}.


\section{The Challenge}
\label{sec:challenge}

    The objectives of the NTIRE 2020 challenge on single image dehazing are: (i) to gauge and push the state-of-the-art in image dehazing; (ii) to compare different solutions; and (iii) to promote the first non-homogeneous dehazing dataset (\textbf{NH-Haze}~\cite{ancuti2020NH-HAZE}) that contains real non-homogeneous haze and ground truth haze-free images.
    
    \subsection{NH-Haze Dataset}
    
    \textbf{NH-Haze}~\cite{ancuti2020ntire} dataset contains 55 various outdoor scenes captured in presence or absence of haze.  \textbf{NH-Haze} is the first dehazing dataset that contains nonhomogeneous hazy scenes. Our dataset allows to investigate the contribution of the haze over the scene visibility by analyzing the scene objects radiance starting from the camera proximity to a maximum distance of 20-30m.

    The recording outdoor conditions had to be similar to the ones encountered in hazy days and therefore, the recording period has been spread over more than two months during the autumn season. Basically, all outdoor scenes have been recorded during cloudy days, in the morning or in the sunset. We also had to deal with the wind speed. In order to limit fast spreading of the haze in the scene, the wind during recording had to be below 2-3 km/h. The absence of wind was the parameter that was the hardest to meet, and explain the long interval of recording the dataset.

   The hardware used to record the scenes was composed of a tripod and a Sony A5000 camera that was remotely controlled (Sony RM-VPR1). We acquired JPG and ARW (RAW) 5456$\times$3632 images, with 24 bit depth. Each scene acquisition has started with a manual adjustment of the camera settings. The shutter-speed (exposure-time), the aperture (F-stop),  the ISO and white-balance parameters have been set at the same level when capturing the haze-free and hazy scene. 
   
   To set the camera parameters (aperture-exposure-ISO), we used an external exponometer (Sekonic) while for setting the  white-balance, we used the middle gray card (18\% gray) of the color checker. For this step we changed the  camera white-balance mode in manual mode and placed the reference grey-card in the front of it. 

   To introduce haze in the outdoor scenes we have used two professional haze machines (LSM1500 PRO 1500 W), which generate vapor particles with diameter size (typically 1 - 10 microns) similar to the particles of the atmospheric haze. The haze machines use cast or platen type aluminum heat exchangers to induce liquid evaporation. We have chosen special (haze) liquid with higher density in order to simulate the effect occurring with water haze over larger distances than the investigated 20-30 meters.
   
   The generation of haze took approximately 2-3 minutes. After starting to generate haze, we used a fan to spread the haze in the scene in order to reach a nonuniform distribution of the  haze in a rage of 20-30 m in front of the camera.

    Moreover, in each outdoor recorded scene  we have placed a color checker (Macbeth color checker) to allow for post-processing. We use a classical Macbeth color checker with the size 11 by 8.25 inches with 24 squares of painted samples (4$\times$6 grid).

    \subsection{NonHomogeneous Haze Challenge}
    
    For the NTIRE 2020 dehazing challenge we created a Codalab competition. To access the data and submit their dehazed image results to the CodaLab evaluation server each participant had to register. 
    
    \textbf{Challenge phases:} (1) Development (training) phase: the participants got train data (hazy and haze-free images) (45 sets of images); (2) Validation phase: the participants received 5 additional sets of images and had the opportunity to test their solutions on the hazy validation images and to receive immediate feedback by uploading their results to the server. A validation leader-board is available; (3) Final evaluation (test) phase: the participants got the hazy test images (5 additional set of images) and had to submit both their dehazed images and a description of their methods before the challenge deadline. One week later the final results were made available to the participants. 
    
    \textbf{Evaluation protocol:} The Peak Signal-to-Noise Ratio (PSNR) measured in decibel (dB) and the Structural Similarity index (SSIM)  computed between an image result and the ground truth are the quantitative measures. The higher the score is, the better the restoration fidelity to the ground truth image is. Additionally we used the perceptual measures LPIPS~\cite{zhang2018perceptual} and Perceptual Index (PI)~\cite{Blau_2018_ECCV_Workshops}, recently used for assessing the quality of the super-resolved images. The final ranking is based on a user study and Mean Opinion Scores (MOS). 
    
\begin{table*}[ht!]
    \centering
    \footnotesize
    \resizebox{\linewidth}{!}
    {
    \begin{tabular}{ll||ll|llr||rccccc}
    \multicolumn{2}{c||}{\large Participant} & \multicolumn{5}{c||}{\large Results} &\multicolumn{6}{c}{\large Solution details}\\
    &&\multicolumn{2}{c|}{Fidelity}&\multicolumn{3}{c||}{Perceptual quality}&Runtime&GPU/&extra&&deep learning&loss\\
    Team & User & PSNR$\uparrow$ & SSIM$\uparrow$ & LPIPS$\downarrow$ & PI$\downarrow$ & MOS$\downarrow$&  img.[s] & CPU & data & ens. & framework&\\
    \hline\hline

\\
\multicolumn{13}{c}{\large Top perceptual quality solutions}\\
\hline
ECNU-Trident  & liujing1995 & 21.41$_{(3)}$ & \textbf{0.71}$_{(1)}$ & \textbf{0.267}$_{(1)}$ & 3.063 & 1 & 8$\times$0.64 & 1080ti  & DenseHaze & 8$\times$ & PyTorch &L1,FFT,BReLU\\ 
ECNU-KT  & asakusa/glassy & 20.85$_{(4)}$ & 0.69$_{(2)}$ & 0.285$_{(2)}$ & 3.295 & 2 & 0.30 & 2080ti & DenseHaze &   & PyTorch & L1, Lap, KT \\

dehaze\_sneaker  & dehazing\_sneaker & 21.60$_{(2)}$ & 0.67 & 0.363 & 3.712 & 3 & 0.21 & v100 & DenseHaze &  & PyTorch &L1,L2 \\
Spider & spider & \textbf{21.91}$_{(1)}$ & 0.69$_{(2)}$ & 0.361 & 3.700 & 4 & 0.22 & v100 & DenseHaze & & PyTorch &L1,L2 \\
NTU-Dehazing & peter980421 & 20.11$_{(5)}$ & 0.66 & 0.351 & 3.973 & 5 & 8.00 & 1080ti & DenseHaze,O-Haze &  &PyTorch \\

VICLAB-DoNET & DoNET & 19.70 & 0.68$_{(3)}$ & 0.301 & 2.985 & 6 & 8$\times$0.95 & Titan XP & DenseHaze,O-Haze & 8$\times$ & PyTorch & L1\\
iPAL-NonLocal  & krm & 20.10 & 0.69$_{(2)}$ & 0.330 & 3.278 & 7 & 2.06 & Titan XP & DenseHaze,O-Haze &  & PyTorch & L1 \\

Team JJ  & pushthebell & 19.49 & 0.66 & 0.311 & 2.824$_{(2)}$ & 8 & 2.38 & 1070 & n/a   & &PyTorch & L1 \\
VIP\_UNIST & Eun-Sung & 18.77 & 0.54 & 0.525 & 4.374 &  9 & 0.04 & n/a  & n/a  &  &  \\
iPAL-EDN & venkat2 & 19.22 & 0.66 & \textbf{0.266}$_{(1)}$ & 3.267 & 10 & n/a &Titan XP & DenseHaze,O-Haze &  & PyTorch & L1\\
\hline

\hline
\\
\multicolumn{13}{c}{\large Medium perceptual quality solutions}\\
\hline
NTUEE\_LINLAB & NTUEE\_LINLAB & 19.25 & 0.60 & 0.426 & 5.061 & 12 & 12.88 & v100 & n/a &  &PyTorch\\ 
NTUST\_merg & aes & 17.74 & 0.63 & 0.322 & 2.899 & 12 & 1.10 &  v100 & n/a &  &PyTorch\\
iPAL-EDN  & wechat & 18.58 & 0.63 & 0.303 & 3.323 & 12 & 0.08  &Titan XP & DenseHaze,O-Haze &  & PyTorch & L1 \\
SIAT & weilan & 17.95 & 0.63 & 0.332 & 3.046 & 12 & 0.03 & 1080ti & n/a &  & \\
HRDN  & xiqi & 18.51 & 0.68$_{(3)}$ & 0.308 & 2.988 & 12 & 13.00 & n/a & n/a &  &PyTorch \\
iPAL-EDN  & yu.2359 & 19.76 & 0.67 & 0.289$_{(3)}$ & 3.535 & 12 & 0.09 &Titan XP & DenseHaze,O-Haze &  & PyTorch & L1 \\

\hline
\\
\multicolumn{13}{c}{\large Low perceptual quality solutions}\\
\hline
Neptune & neptuneai & 17.77 & 0.62 & 0.407 & 3.413 & 14 & 2.80 & n/a & n/a &  &PyTorch \\
sinashish  & sinashish & 17.11 & 0.62 & 0.357 & 3.141 & 14 & 11.30 & n/a & n/a &  & \\
Neuro\_avengers & souryadipta & 18.24 & 0.65 & 0.329 & 3.051 & 14 & 0.01 & 1080 & no &  &PyTorch & Hybrid Loss\\ 
NITREXZ  & xuanzhao & 18.70 & 0.64 & 0.328 & 3.114 & 14 & 10.43 & 1080ti & n/a &  &PyTorch \\

\hline
\\
\multicolumn{13}{c}{\large Lowest perceptual quality solutions}\\
\hline
IIT\_ISM  & ayu\_22 & 15.29 & 0.57 & 0.457 & 4.451 & 16 & 0.05 & n/a & n/a &  & \\
AISAIL & bouyang & 18.67 & 0.64 & 0.303 & 3.211 & 16  & 1.64 & TitanXP & n/a &  &TensorFlow \\
ICAIS\_Dehaze  & chongya & 16.87 & 0.58 & 0.428 & 2.959 & 16  & v100 & n/a & n/a &  &PyTorch \\
RETINA & de20ce & 12.80 & 0.41 & 0.534 & 3.097 & 16  & 600.00 & n/a & n/a &  & \\
CVML & specialre & 17.88 & 0.57 & 0.378 & 2.855$_{(3)}$ & 16  & 0.06 & n/a & n/a &  & \\
FAU Harbour Branch & destrada2013 & 18.67 & 0.64 & 0.303 & 3.211 & 16  & 13.21 & n/a & n/a &  & \\
hazefreeworld & Navaneeth\_R & 15.88 & 0.38 & 0.766 & 10.676 & 16  & 0.95 & Titan V & O\&I-Haze,HazeRD,DHazy & & PyTorch & Hybrid Loss\\ 

    \hline
    no processing & \textit{baseline}& 11.33 & 0.42 &0.582& \textbf{2.609}$_{(1)}$&20&\\
\hline
    \end{tabular}
    }
	\caption{NTIRE 2020 NonHomogeneous Dehazing Challenge preliminary results in terms of PSNR, SSIM, LPIPS~\cite{zhang2018perceptual}, PI~\cite{Blau_2018_ECCV_Workshops} and Mean Opinion Score (MOS), on the NH-Haze test data. 
	}
	\label{table:results}
\end{table*}

\section{Challenge Results}
\label{sec:challenge_results}

\begin{figure}[!htb]
		\centering
		\includegraphics[width=1\linewidth]{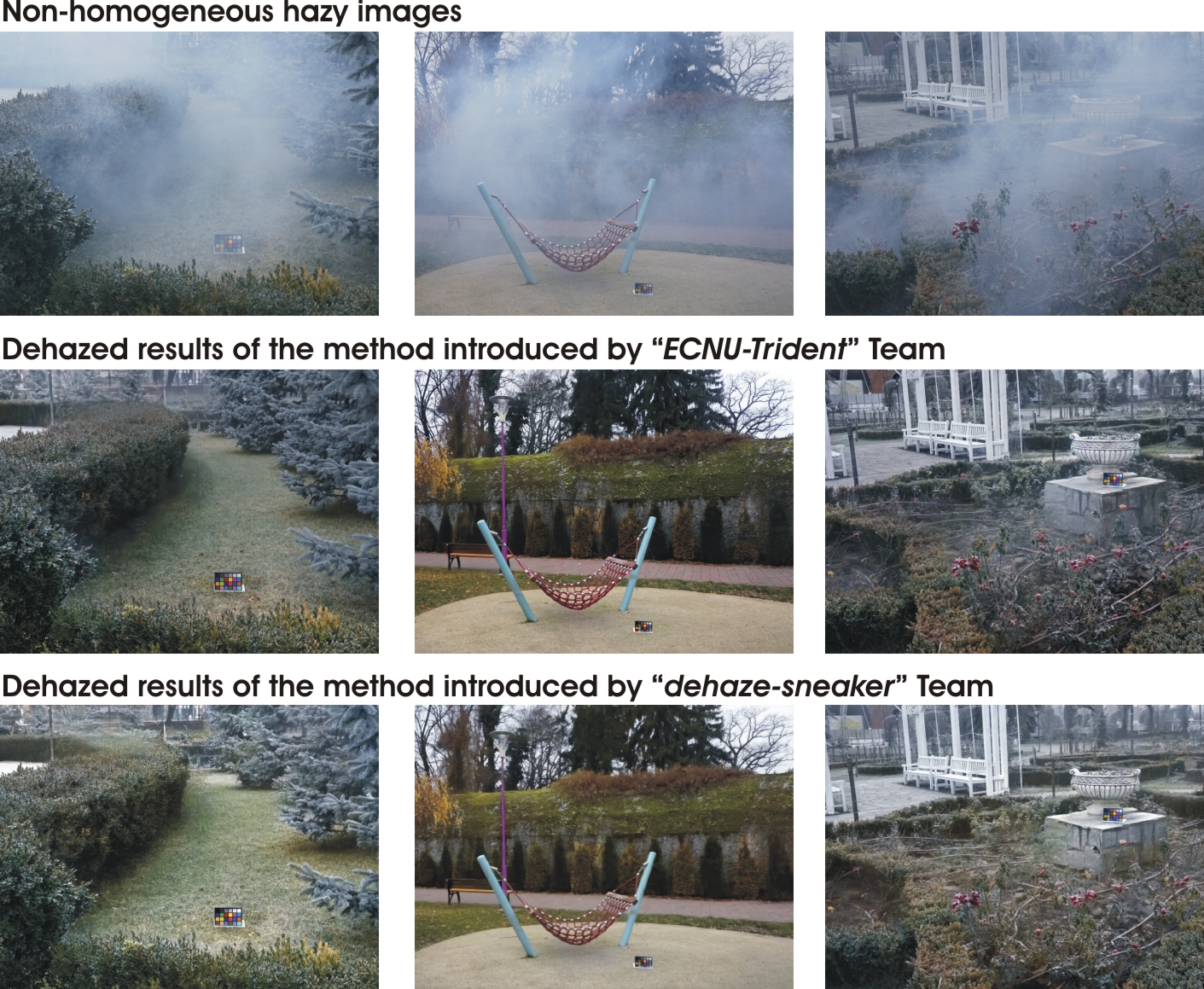}\\
		\caption{Comparative results. The first row displays 3 input hazy images of the NH-HAZE dataset. The second row shows the results yielded by the ECNU-Trident team (winner of NTIRE 2020 dehazing challenge). The bottom row displays the results yielded by the dehaze-sneaker team (3rd in the NTIRE 2020 dehazing challenge).}
		\label{fig:results}
	\end{figure}

From 168 registered participants, 27 teams were ranked in the final phase. These teams submitted results, codes, and factsheets. The fidelity and perceptual quality quantitative results and the final perceptual MOS-based ranking of the challenge are reported in Table~\ref{table:results}. Note that for completeness we report also results of some other submissions that were not ranked due to various reasons such as incomplete submissions, intermediary results, etc.
Figure~\ref{fig:results} allow a visual inspection of the dehazing results obtained by a selection of methods.

From Table~\ref{table:results} and Figure~\ref{fig:results} we can make the following observations. First, ECNU-Trident is the winner of the challenge, ECNU\_KT and dehaze\_sneaker win the second and third place, respectively. Second, among the top perctual quality solutions, dehaze\_sneaker and Spider achieve a good trade-off between fidelity and perceptual quality and runtime requirements. Third, Spider achieves the best fidelity in terms of PSNR, however it fails to provide also the best perceptual quality. Fourth, PSNR and SSIM were better indicators of the perceptual quality of the results, than the LPIPS (full reference)~\cite{zhang2018unreasonable} and PI (no reference)~\cite{Blau_2018_ECCV_Workshops} perceptual quality assessment measures. 
The top-5 PSNR scored teams are also in the top-5 perceptual rank, while the team achieving the best LPIPS score was ranked $10^{th}$ by the MOS, and the best PI score was produced by the team credited with the $8^{th}$ place. Fifth, even for the best solutions it is easy to distinguish the ground-truth haze-free images from the dehazed ones.\\ 
\noindent\textbf{Architectures, losses and main ideas }
Many of the proposed solutions employ and are inspired from architectures such as U-Net~\cite{ronneberger2015unet}, ResNet~\cite{he2016deepresidual}, DenseNet~\cite{DenseNet} and Inception~\cite{szegedy2017inception}. The winner ECNU-Trident combines three different sub-networks. ECNU-KT proposes a dual network made from a teacher and a dehazing pair of networks. We note also that top ranked teams such as dehaze\_sneaker and Spider use self-designed Haze-Aware Representation Distillation (HARD) modules.
L1 is the most employed loss for training the deep learned dehazing networks, however combinations are also found. ECNU-Trident employs L1, FFT, BReLU losses. ECNU-KT employs L1, Laplacian (Lap), and Knowledge Transfer (KT) losses. dehaze\_sneaker and Spider rely and L1 and L2 losses, while Neuro\_avengers uses a hybrid loss.

\noindent\textbf{Ensembles }
Many teams, including the winners, employed commonly used model-ensemble or self-ensemble~\cite{Timofte_2016_CVPR} to improve the performance of their solutions.\\
\noindent\textbf{Train data}
Most of the top ranked teams used the provided NH-Haze train data and augmented the training data with data from DenseHaze~\cite{Ancuti_DENSE_HAZE_2019} and O-Haze~\cite{Ancuti_OHAZE_2018}, datasets providing pairs of real hazy and haze-free images. Some teams use pretrained models on other datasets/tasks.\\
\noindent\textbf{Deep learning platforms } The vast majority of the proposed solutions use PyTorch platform, while a couple use other such as TensorFlow.\\
\noindent\textbf{Runtime } The self-reported runtimes per processed image range from 0.01s (Neuro-avengers) on a Nvidia GTX 1080 GPU card to 600s (RETINA) on a CPU.\\  
\noindent\textbf{Conclusions}
By analyzing the challenge methods and their
results we can draw several conclusions. (i) The proposed
solutions have a degree of novelty and go beyond the published
state-of-the-art methods. (ii) In general the best perceptual quality solutions
performed the best also for both PSNR and SSIM fidelity measures. (iii) The evaluation based on the perceptual measures (LPIPS and PI) is questionable since these measures were not tailored for hazy scenes.

\section{Challenge Methods and Teams}

\subsection{ECNU-Trident}

\begin{figure*}[t]
	\centering
	\includegraphics[width=17.6cm]{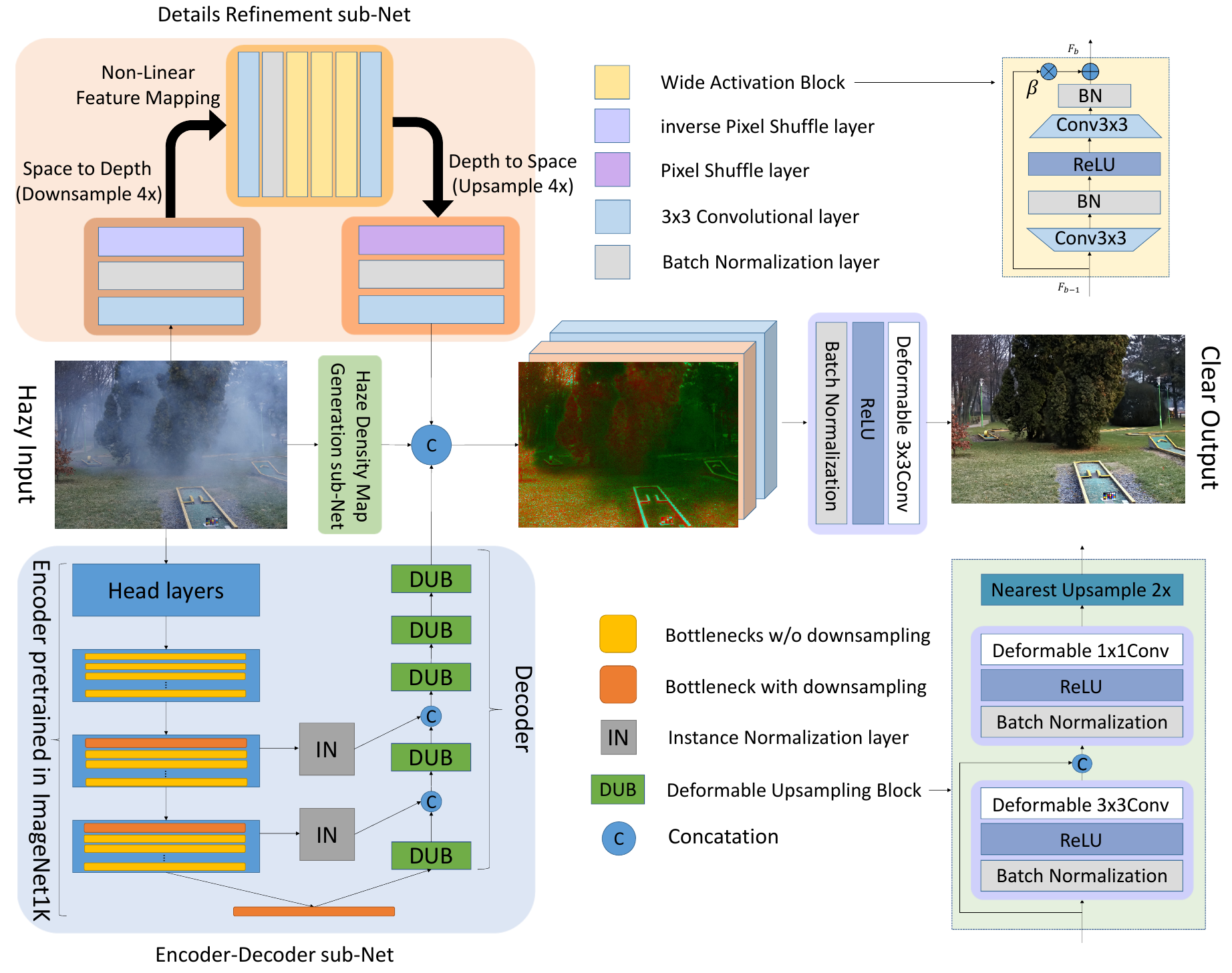}
	\caption{The architecture of the proposed Trident Dehazing Network (TDN), the Details Refinement sub-Net (DRN) and the Encoder-Decoder sub-Net (EDN). $\oplus$ represents tensor addition and $\otimes$ represents tensor multiplication respectively. TDN consists of three sub-nets: EDN, DRN and HDMGN. The haze density maps and intermediate feature maps output by three sub-nets are then concatenated and fed into the tail deformable \cite{zhu2019deformable} convolution block to get the clear output.}

	\label{fig2net1}
\end{figure*}

\begin{figure}[t]
	\centering
	\includegraphics[width=8.4cm]{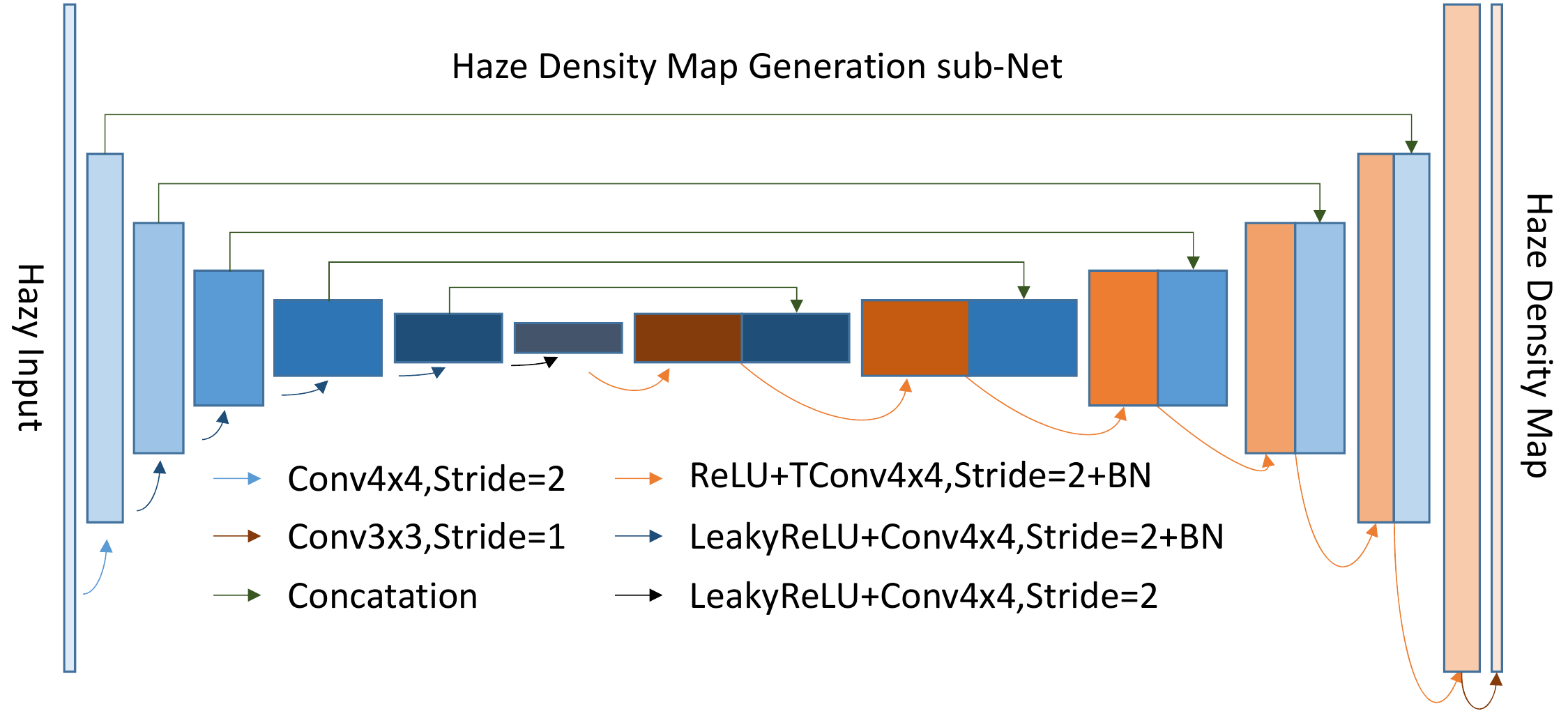}
	\caption{The architecture of the proposed Haze Density Map Generation sub-Net. ``TConv" is the abbreviation of ``Transpose Convolution".}
	\label{fig3unet}
\end{figure}

ECNU-Trident team proposed a Trident Dehazing Network (TDN)~\cite{liu2020trident} to directly learn a mapping from the input real world nonhomogeneous hazy image to the hazy-free clear image. As shown in Figure~\ref{fig2net1}, TDN consists of three sub-nets, the Encoder-Decoder sub-Net (EDN), the Details Refinement sub-Net (DRN), and the Haze Density Map Generation sub-Net (HDMGN), each of which is used for a specific purpose: EDN reconstructs the coarse features of hazy-free images, DRN complements the high frequency details of the hazy free image features, and HDMGN helps obtaining the density of haze in the different region of the input hazy image. The deformable~\cite{zhu2019deformable} convolution block gets the final clear output from the concatenated feature maps of three sub-nets.

DPN92 pretrained in ImageNet1K is as the backbone of EDN's encoder part. The decoder is composed of five Deformable Upsampling Blocks (DUB), as shown in Figure 1 (bottom right). The input feature is first fed into a 3$\times$3 deformable convolution block, and then concatenated with the output features. The concatenated features are fed into an 1$\times$1 deformable convolution block and an nearest-upsampling 2x layer to get the upsampled features as the input features of the next DUB. EDN adds skip connections from the output of the first downsampling block in layer 2 and that in layer 3 to the input of DUB 2, 3 by concatenating (cat) the feature maps, respectively. EDN use trainable instance normalization \cite{ulyanov2016instance} for skip connections.

As shown in Figure \ref{fig3unet}, HDMGN is a U-Net architecture proposed in pix2pix \cite{isola2017image} network to achieve haze density map generation. Different with U-Net in pix2pix network, HDMGN adds a tail 3$\times$3 convolutional layer to refine the output. Due to the size division requirement, there are only 6 downsampling and upsampling operators in the U-Net, and the input size should be divisible by 64. As shown in Figure \ref{fig2net1}, the greener the region in the visualization haze density map is, the more haze there is.

DRN does the nonlinear feature mapping on downsampled 4$\times$factor. Inverse Pixel Shuffle layer is used to change the feature maps from spatial to depth (downsampling/desubpixel), and Pixel Shuffle layer \cite{shi2016real} is used to change the feature maps from depth to spatial (upsampling/subpixel). As shown in Figure \ref{fig2net1}, three Wide Activation Blocks (WAB) provide the non-linear feature mapping on 4$\times$ downsampled factor. In the WAB, there are two 3 $\times$3 convolutional layers (followed by batch normalization layer) and a wide activation layer proposed in \cite{yu2018wide}. The channel expand factor of WAB is 4. Motivated by \cite{wang2018esrgan}, ECNU-Trident uses residual scaling, \ie, scaling down the residuals by multiplying a constant between 0 and 1 before adding them to the main path, to prevent training-instability.

\subsection{ECNU-KT}

\begin{figure*}[!htb]
		\centering
		\includegraphics[width=1\linewidth]{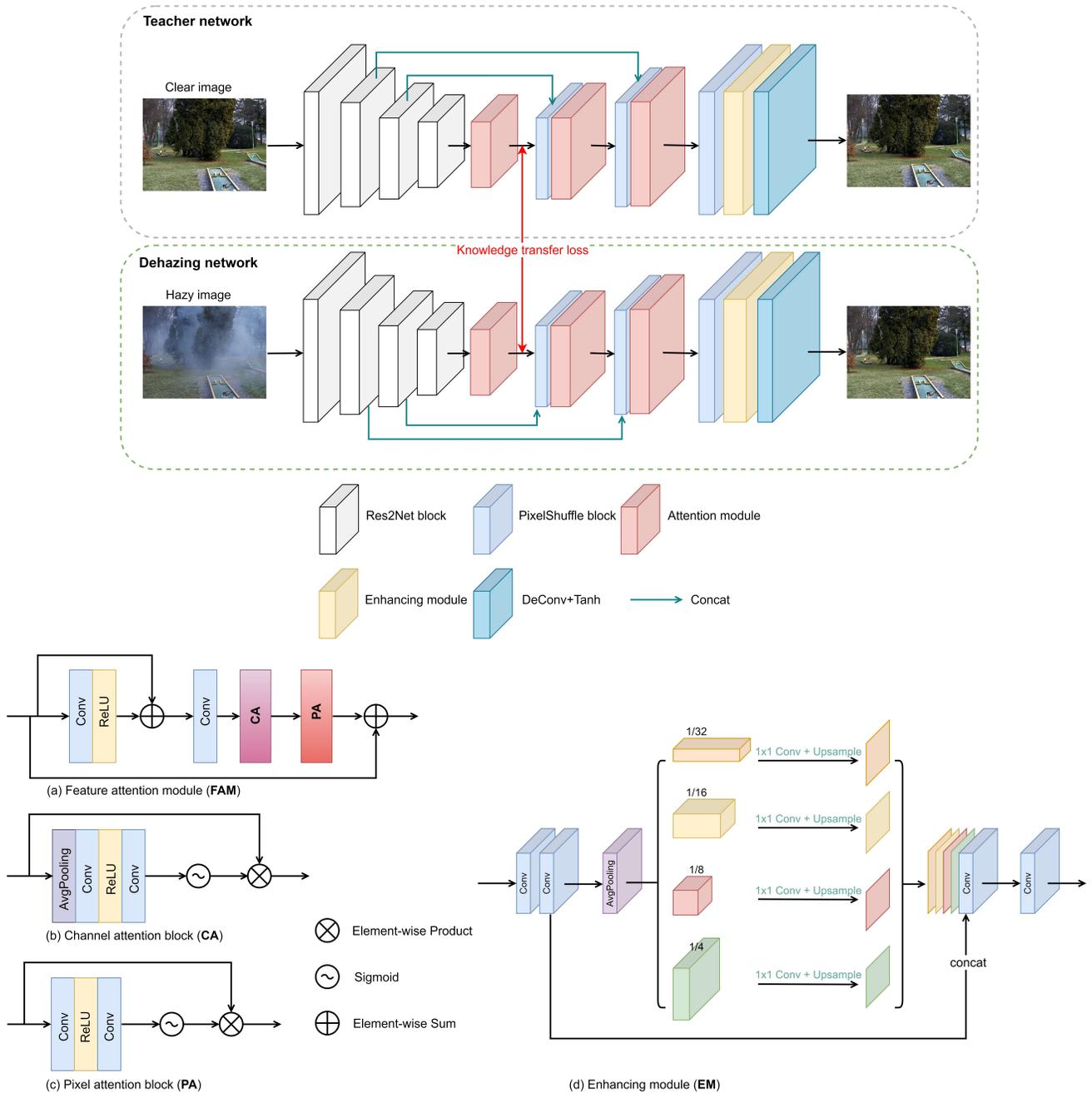}\\
		\caption{Overview of the KTDN architecture. }
		\label{fig:KTDN}
\end{figure*}

The ECNU-KT team proposed a knowledge transfer method~\cite{wu2020knowledge} that utilizes abundant clear images to train a teacher network which can learn strong and robust prior. It supervises the intermediate features and uses the feature similarity to encourage the dehazing network imitate the teacher network. The prior knowledge are transferred to the dehazing network by intermediate feature map. The method is based on a dual network that consists of the teacher network and the dehazing network, as shown in Figure~\ref{fig:KTDN}. The architectures of networks are identical and both are based on encoder-decoder structure. In addition, the method uses a pre-trained  Res2Net~\cite{gao2019res2net} without FC layer and only downsample 16x as encoder to extract detail information of hazy images, and add skip connection to preserve information. Moreover, in order to process nonhomogeneous hazy images, inspired by~\cite{qin2019ffa}, the method  uses the feature attention module (FAM) that combines channel attention with pixel attention to let network pay more attention to effective information such as texture, color and thick haze region. Finally, the method has  an enhancing module (EM) to refine results inspired by \cite{qu2019enhanced}.
\\ The teacher network is trained first, followed by the dehazing network.

\subsection{dehaze\_sneaker}

\begin{figure}
\centering
\includegraphics[width=1\linewidth]{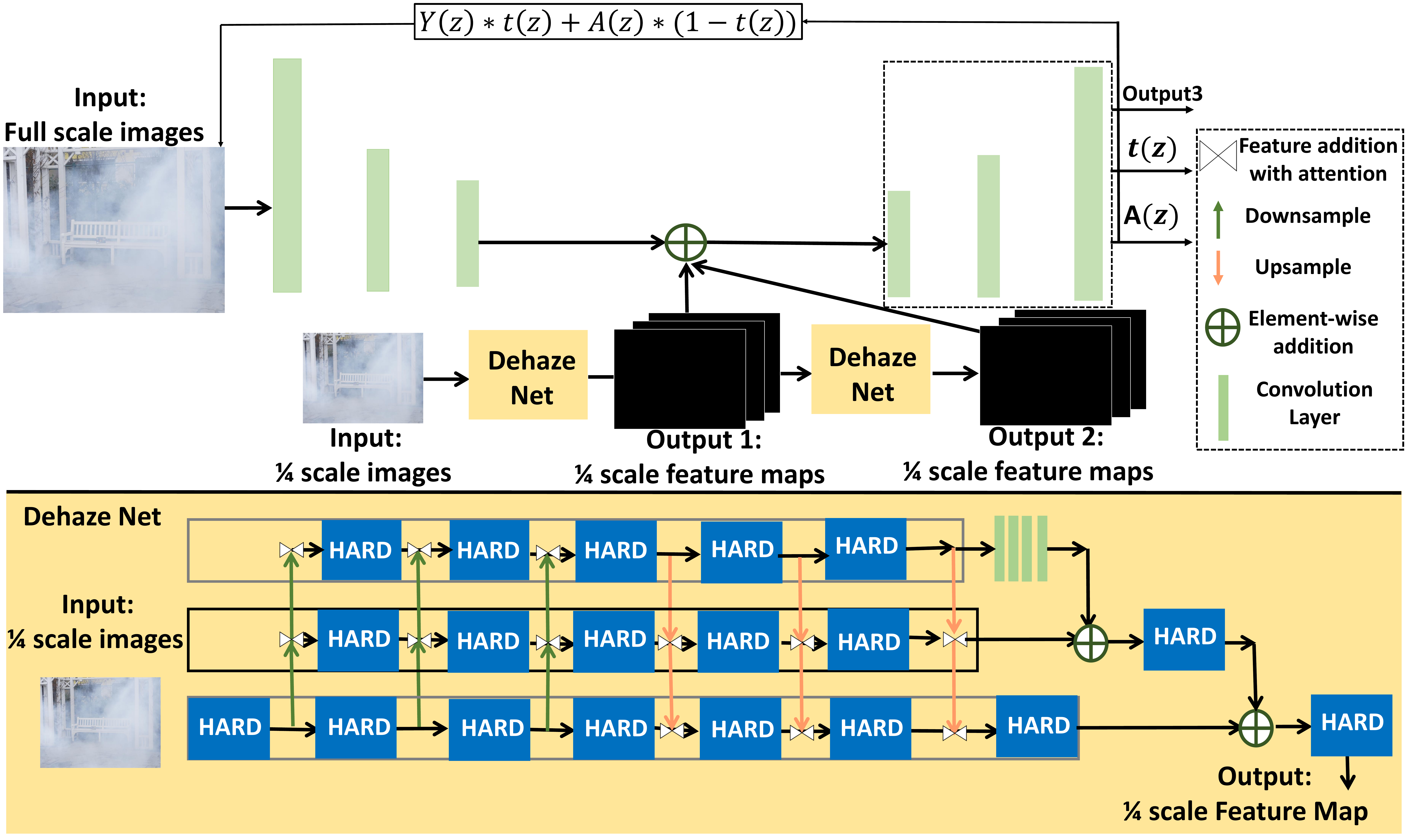}
\caption{Overview of the framework introduced by \textbf{dehaze\_sneaker} team.  It contains one encode-decoder framework for full scale images processing, while DehazeNet is proposed to handle $\frac{1}{4}$ scale images to provide haze-free information to framework. In the bottom are shows the details for DehazeNet. $\frac{1}{4}$ scale images are feed into the DehazeNet.}
\label{fig:framework}
\end{figure}

The proposed framework is illustrated in Figure~\ref{fig:framework}. The main framework acts like an encoder-decoder. In the center of framework, DehazeNet is proposed to provide haze-free information to final results.

As illustrated in Fig.~\ref{fig:framework}, the DehazeNet consists of three layers from coarse to fine: the first (coarsest) layer involving five Haze-Aware Representation Distillation (HARD) modules, the second (medium) with five HARDs, and the third (finest) with seven HARDs. Given an input hazy image $X$ and its target haze-free image $Y$, let $x_{m}^{n}$ and $y_{m}^{n}$ denote the input and output of the $n$-th HARD in  the $m$-th layer. 

The DehazeNet at each scale starts from the finest scale($1/4$) and sequentially passes the extracted features up to the coarsest ($1/16$) scale. Then the information will be sent back to finest scale($1/4$) as haze-free information.

\begin{figure}[!htb]
		\centering
		\includegraphics[width=1\linewidth]{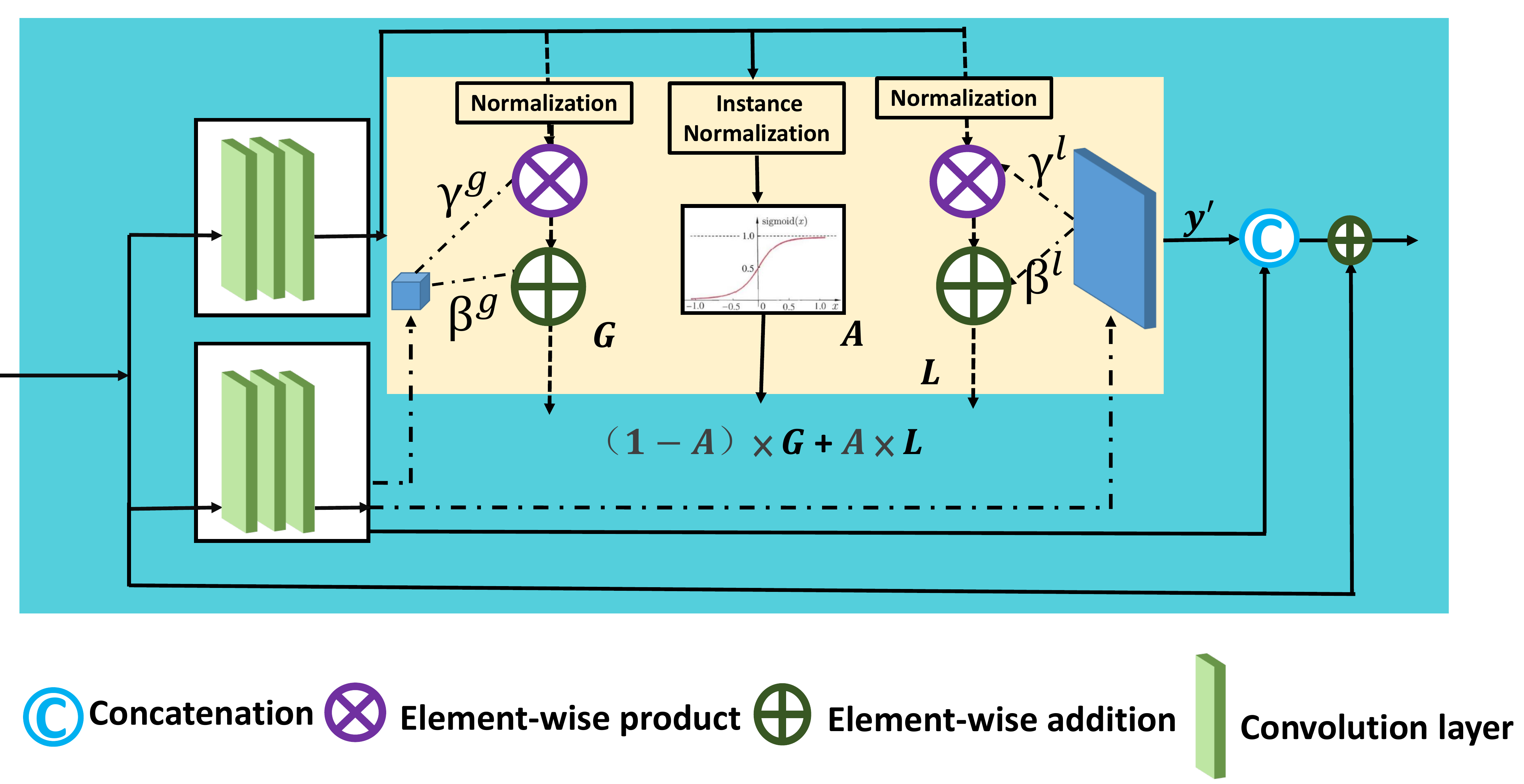}\\
		\caption{\textbf{Haze-Aware Representation Distillation (HARD) Module.} HARD is composed of two branches. The second branch is used to learn spatial information $\gamma_g$, $\beta_g$ and global atmospheric light information $\gamma_l$, $\beta_l$, then feed them into the first branch to form intermediate results $y'$. After channel attention, the final result of HARD is produced.}
		\label{fig:dh_module}
	\end{figure}

The HARD Module contains two branches, as shown in Fig. \ref{fig:dh_module}. Because haze in the real world is always in an irregular pattern, the attention map is proposed to combine atmospheric light and spatial information together selectively. \textbf{In this method, we calculate the final result by mathematical formulation proposed by \cite{mccartney1976optics}.}

\subsection{Spider}

The proposed framework is illustrated in Figure~\ref{fig:framework}. The main framework acts like an encoder-decoder. In the center of framework, DehazeNet is proposed to provide haze-free information to final results.

\begin{figure}
\centering
\includegraphics[width=1\linewidth]{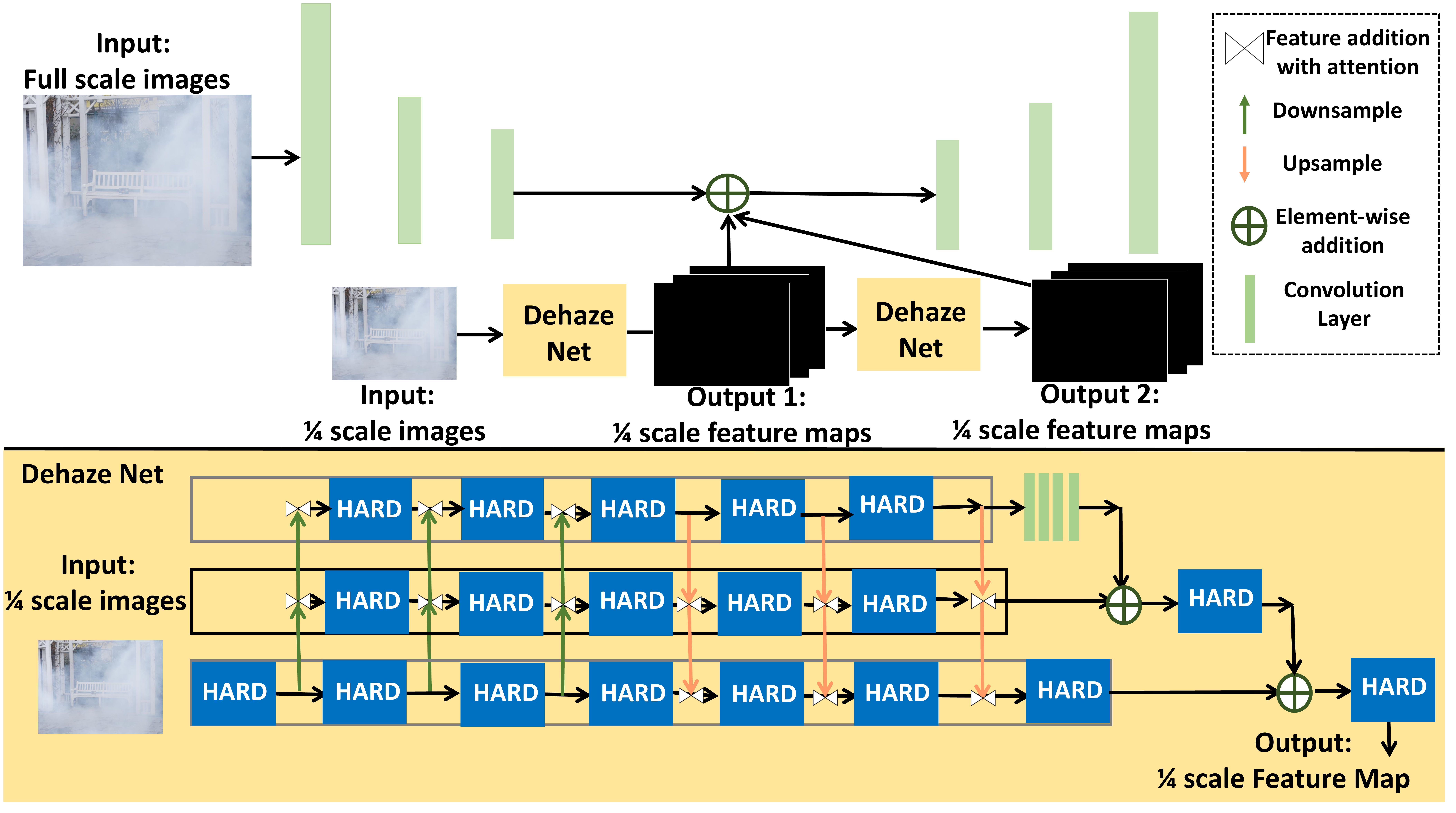}
\caption{Overview of the method proposed by Spider team. The upper is the overview for our proposed methods. It contains one encode-decoder framework for full scale images processing, while DehazeNet is proposed to handle $\frac{1}{4}$ scale images to provide haze-free information to framework. The lower one shows details for DehazeNet. $\frac{1}{4}$ scale images are feed into the DehazeNet.}
\label{fig:framework2}
\end{figure}

As illustrated in Fig. \ref{fig:framework2}, the DehazeNet consists of three layers from coarse to fine: the first (coarsest) layer involving five Haze-Aware Representation Distillation (HARD) modules, the second (medium) with five HARDs, and the third (finest) with seven HARDs. Given an input hazy image $X$ and its target haze-free image $Y$, let $x_{m}^{n}$ and $y_{m}^{n}$ denote the input and output of the $n$-th HARD in  the $m$-th layer. 

The DehazeNet at each scale starts from the finest scale($1/4$) and sequentially passes the extracted features up to the coarsest ($1/16$) scale. Then the information will be sent back to finest scale($1/4$) as haze-free information.

The HARD Module contains two branches, as shown in Fig. \ref{fig:dh_module}. Because haze in the real world is always in an irregular pattern, the attention map is proposed to combine atmospheric light and spatial information together selectively. \textbf{In this method, we output final results directly.}

\subsection{NTU Dehazing}

\begin{figure}
\begin{center}
    \includegraphics[width=1\linewidth]{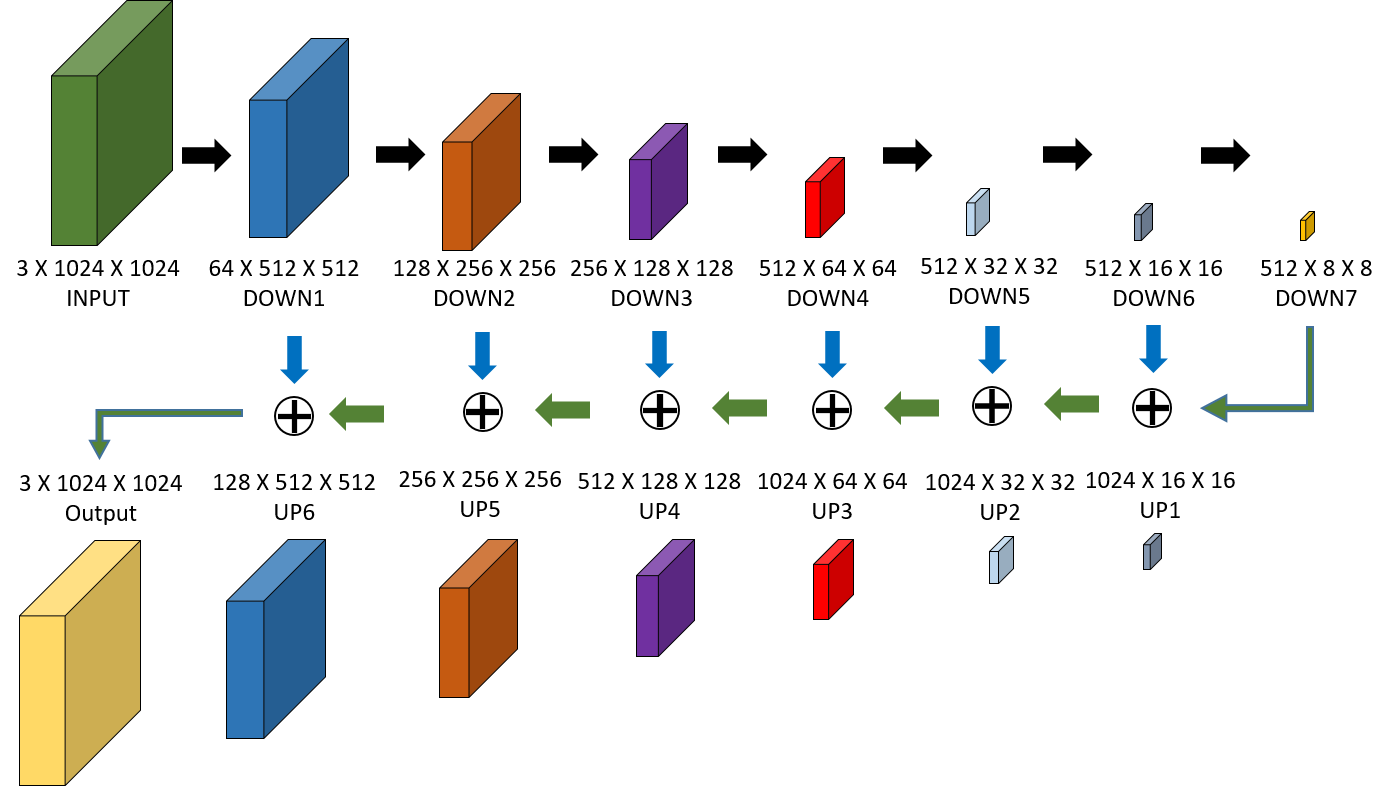}
\caption{Overview of the method introduced by NTU Dehazing team.}
\label{fig:inference process}
\end{center}
\end{figure}

The team members proposed a customized UNet\cite{ronneberger2015unet}, using the residual network\cite{he2016deepresidual} and the Inception module\cite{szegedy2017inception}. Each convolutional layer in the model is followed by instance normalization and LeakyReLU activation with negative slope as $0.2$, except for the last layer of the encoder and the last two layers of the decoder, where the convolutional layers are followed by only activation function. The input and output sizes of the network are both $ 1024 \times 1024 $(High-Resolution). Because of the input and output size, we can get better quality output and avoid distortion caused by downsampling (see Figure~\ref{fig:inference process}).

\subsection{VICLAB-DoNET}

\begin{figure}
    \includegraphics[width=1\linewidth]{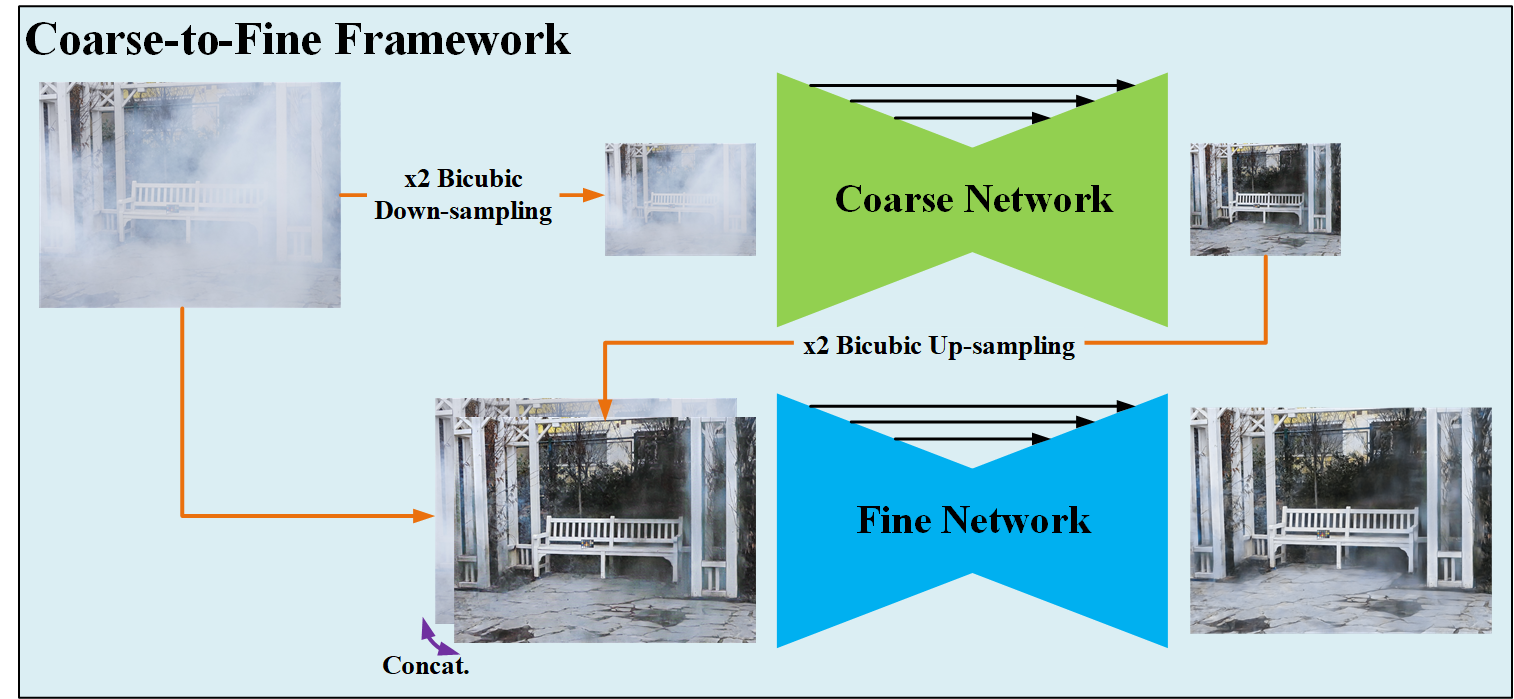}
    \caption{Diagram of the proposed coarse-to-fine framework. The coarse network removes the overall haze, and the fine network reconstructs the colors and details of image. Best viewed in color.}
    \label{fig:figVIC}
\end{figure}

 The authors propose the 2-stage coarse-to-fine framework to remove non-homogeneous haze effectively. This framework consists of coarse network which removes overall haze and fine network which reconstructs colors and details from output of coarse network. The method which uses a large size of kernel and increases the depth of network is able to enlarge the receptive field, but it causes numerous computational complexity.  In other ways, the size of receptive field can be increased by down-sampling the input image through the network. However, if the down-sampled input image is passed through the network and then up-sampled again to acquire the output image without any post-processing, there exists a risk that the output image may be blurry. To solve this problem, the 2-stage coarse-to-fine framework was used. By using down-sampled hazy images as input, non-homogeneous haze can be easily included in receptive field of coarse network, so it effectively removes overall haze. The original hazy image is concatenated with up-sampled output of coarse network, then it  use it as input to the fine network. Hence, the fine network recognizes the hazy parts and restores the colors and details. The detailed structure of the entire framework is shown in Figure \ref{fig:figVIC}.

\subsection{iPAL-NonLocal}

\begin{figure*}
\centering
\includegraphics[width=1\linewidth]{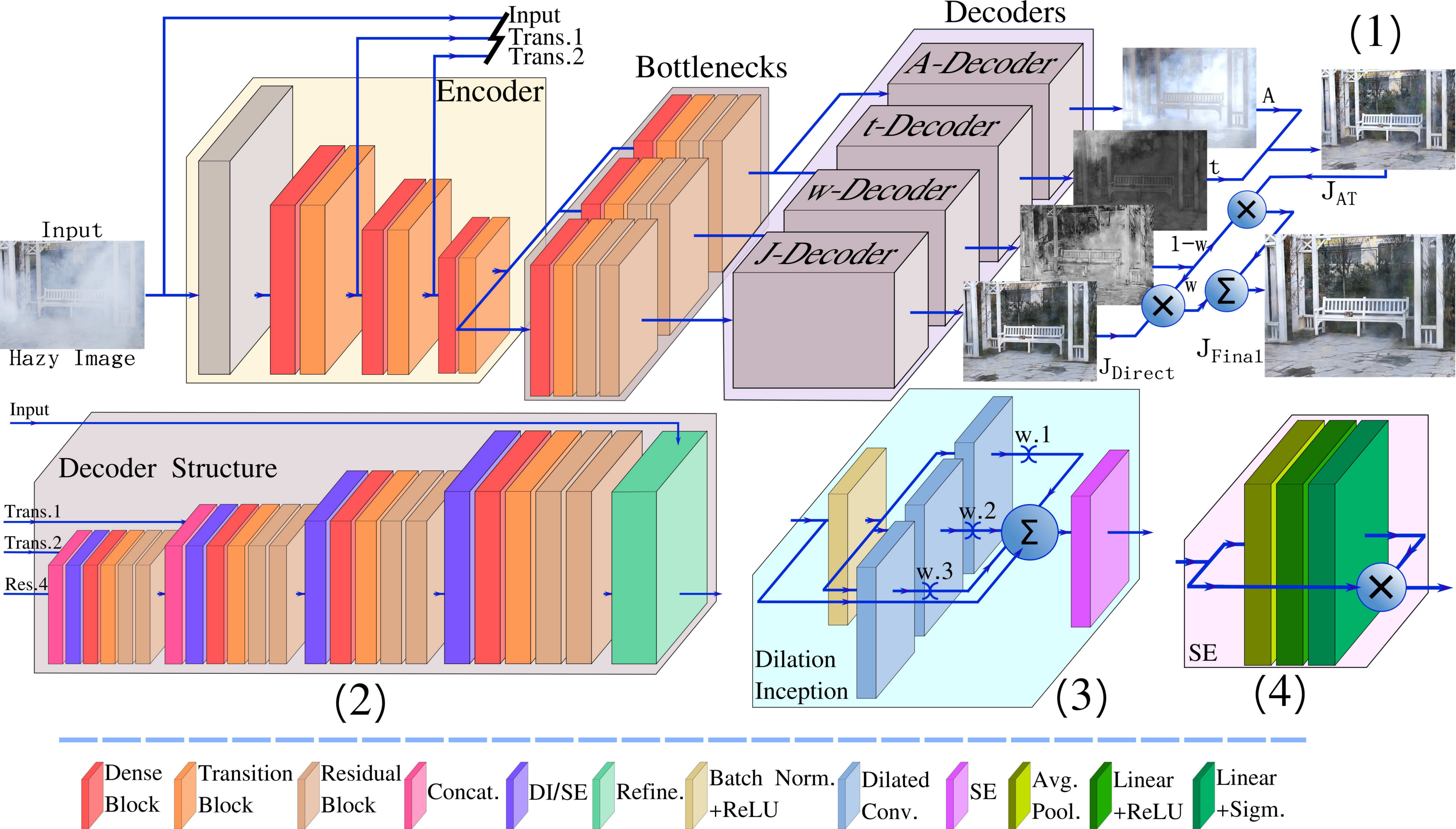}
\caption{Overview of the method introduced by iPAL-NonLocal. All decoders are identical except for $J$-Decoder which has extra layers for non-local features utilization (Dilation Inception--DI). $A$-, $t$- and $w$-decoders only use SE layers.}
\label{fig:atjwd}
\end{figure*}

The method is based on  two proposed models following similar ideology~\cite{Metwaly_NonLocal_2020}, the `AtJw' and the `AtJwD' models. As illustrated in Fig.~\ref{fig:atjwd}, both models have one encoder and four decoders. They share the same architecture in the encoder and three of the decoders but have different architecture in the left decoder --- $J$-Decoder. 

The encoders in both models include three pre-trained dense blocks borrowed from a DenseNet-121 \cite{huang_2017_densely}, which is proposed initially for classification problems. During the training,
in the initial phase, the parameters of the encoder were frozen, and  the parameters of decoders were trained with large training rate. After a certain number of epochs, the parameters of that encoder were optimized by training the whole network together with small learning rate. The intuition behind it is to allow decoders to gain some advantage of the pre-trained dense blocks in the beginning since they are initialized randomly unlike the pre-trained encoder.

The decoders in `AtJw' and `AtJwD' models have similar structures except the $J$-decoders. The first two decoders $t$-decoder and $A$-decoder are responsible for recovering the haze-free image using the dehazing mathematical model.

$J$-decoder is responsible for retrieving the haze free image directly from the input hazy image. Unlike the dehazed image $J_{At}$, $J_\text{direct}$ is generated directly which enables the network to hallucinate regions with very dense haze. Thus for regions where the value of $A$ is high and the value of $t$ is low, $J_\text{direct}$ is expected to perform better than the noisy output of $A+t$-decoders. However, in regions with light-to-no haze, $J_{At}$ will perform better as the direct output lacks sharpness and details. The output of $w$-Decoder, a spatially varying weight map, in testing confirms that conclusion.

\subsection{Team JJ}

\begin{figure}
    \includegraphics[width=1\linewidth]{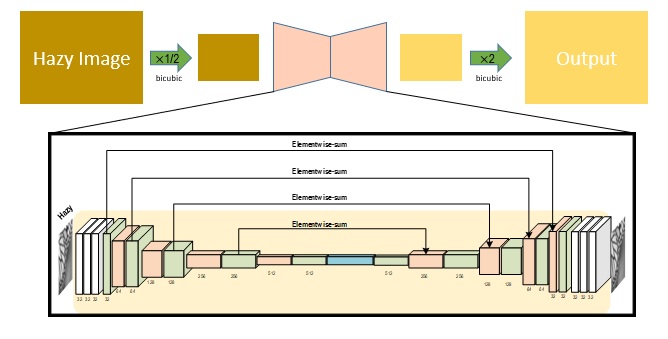}
    \caption{Diagram of the  model introduced by Team JJ.}
    \label{fig:figJJl}
\end{figure}

The authors propose a deep UNet-based model which consists of densenet blocks. For each up/down sampling phase, 
the feature attention block was added in order to sufficiently utilize an important part of the feature map between each up/down blocks. Basically, a fancy bottleneck block was used to compose each densenet block and residual block, which split feature map channels to 4 part and apply convolution differently for each parts so that we consider multi-scale-wise perspective.   The proposed model consists of two main parts, encoder and decoder. Between two parts, there is skip-connection by element-wise summation for two features with same size of feature map. Furthermore, we add residual block between encoder and decoder so that make up the output of encoder to enhance overall performance. Before put images into network model, original image size was decreased by a factor of 1/4 size and the corresponding output image size was increased by factor of 4.The overall architecture is depicted in Figure \ref{fig:figJJl}.

\subsection{iPAL-EDN}

The team  proposes 3 models:` EDN-3J', `EDN-AT' and `EDU' to address the issue of non-homogeneous haze\cite{Yu_ensemble_2020}.

First, the authors propose a DenseNet based single-encoder four-decoders structure denoted as EDN-3J, wherein among the four decoders, three of them output estimates of dehazed images ($\mathbf{J}_1$, $\mathbf{J}_2$, $\mathbf{J}_3$) that are then weighted and combined via weight maps learned by the fourth decoder. In the second model called EDN-AT, the single-encoder four-decoders structure is maintained while three decoders are transformed to jointly estimate two physical inverse haze models that share a common transmission map $\mathbf{t}$ with two distinct ambient light maps ($\mathbf{A}_1,\mathbf{A}_2$). The two inverse haze models are then weighted and combined for the final dehazed image. To endow two sub-models flexibility and to induce capability of modeling non-homogeneous haze, attention masks are applied to ambient lights. Both the weight maps and attention maps are generated from the fourth decoder. Finally, in contrast to the above two ensemble models, an encoder-decoder-U-net structure called EDN-EDU is proposed, which is a sequential hierarchical ensemble of two different dehazing networks with different modeling capacities.\\

\begin{figure}[h!]
\includegraphics[width=\linewidth]{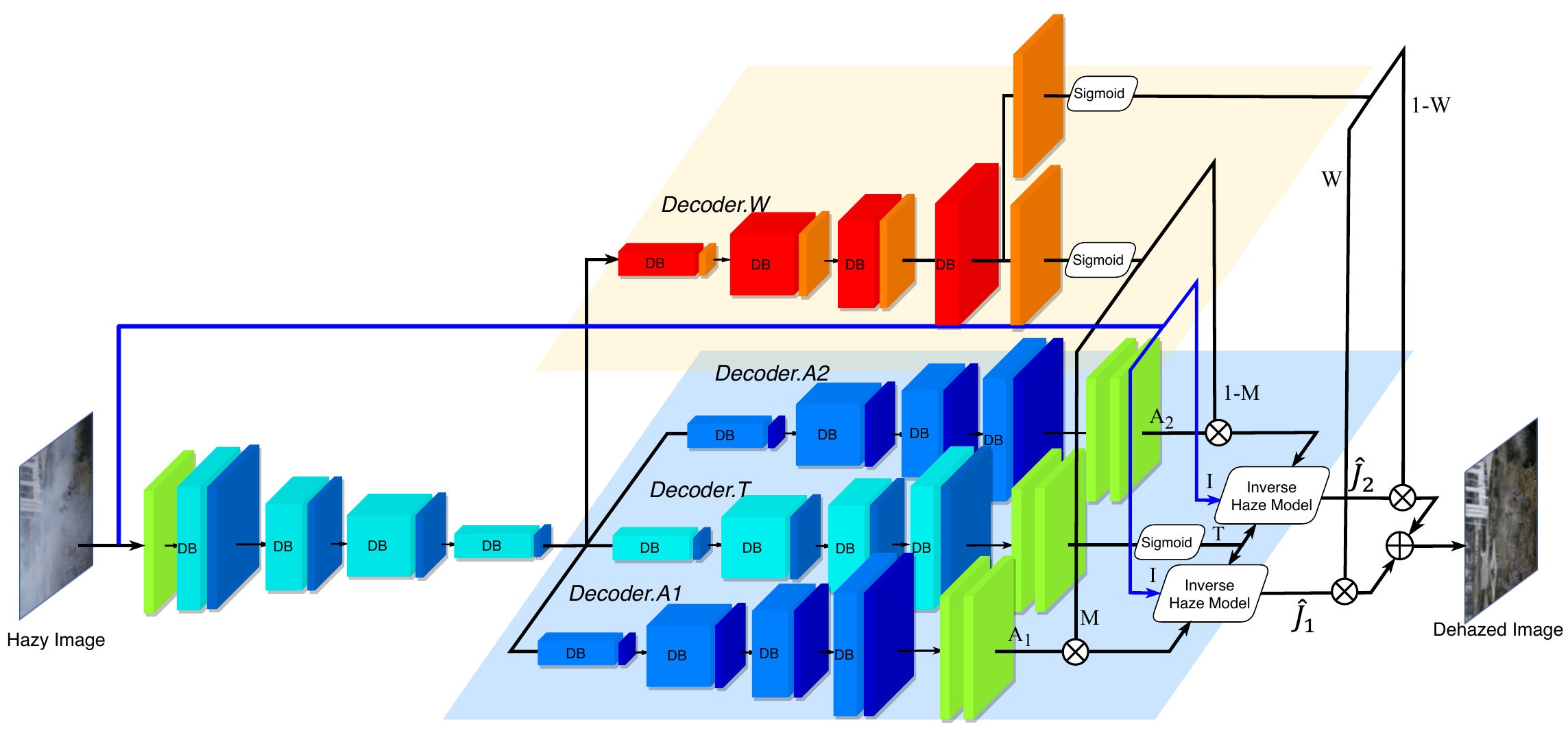}
\caption{Overview of the method proposed by iPAL-EDN team. In EDN-AT model, ambient light maps $\mathbf{A}_1, \mathbf{A}_2$, transmission map $\mathbf{t}$ are estimated by decoder.A1, decoder.A2 and decoder.T respectively. $\mathbf{A}_1, \mathbf{A}_2$ are multiplied by attention maps $\mathbf{m}$ and $1-\mathbf{m}$. Final output is a weighted combination of two sub-models' outputs.}
\label{fig:EDN-AT}
\end{figure}

\subsection{NTUEE\_LINLAB}

The method is based on a encoder-decoder generator model with a multi-scale kernel encoder in the front (size is 3, 5, and 7). It is  trained with part of the densenet-161. Next, the authors used  BicycleGAN to enhance the generator. The hazy image and the difference of the hazy image and ground truth are used as the input when training the cVAE-GAN Encoder.

\subsection{NTUST-merg}

The proposed method is based on the $At-DH$ Network
as the backbone network. The authors used  use DenseNet as the pretrained model in the encoder network. On the other hand, they employed the similar structure of the DenseNet with additional residual block in the decoder network.
They used two decoders to estimate A (atmosphere light) and t(transmission map).  Moreover, L2 loss and perceptual loss were used as loss functions. In the loss term, they both only calculate the estimated haze-free image and ground truth loss.

\subsection{SIAT}

\begin{figure}[h!]
 \centering
 \includegraphics[width=1\linewidth]{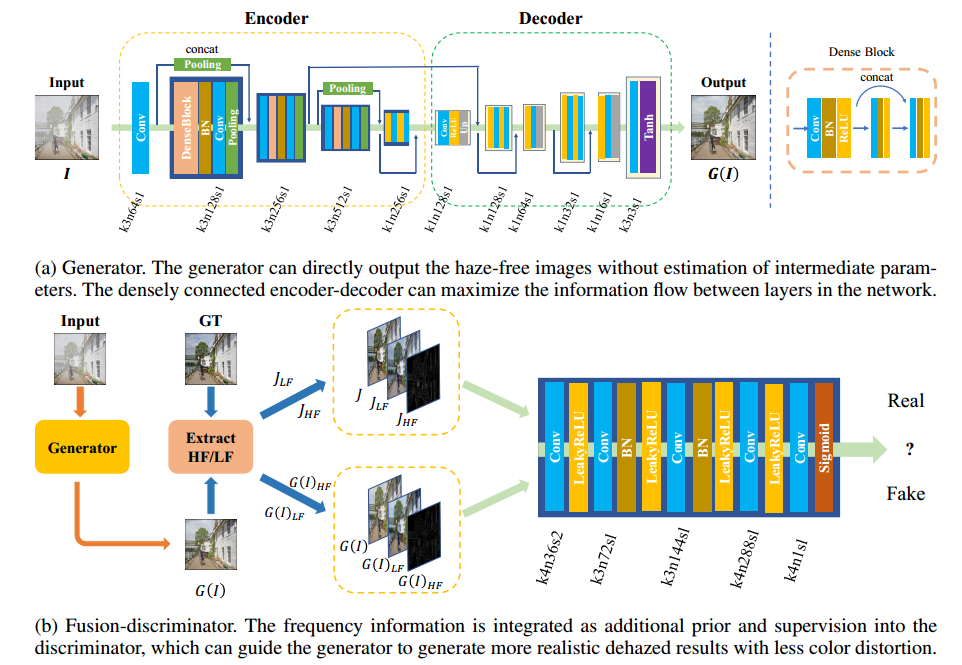}
 \caption{Architecture of the method introduced by SIAT team.}
 \label{fig:SIAT}
 \end{figure}

The proposed method  is a fully end-to-end algorithm for image dehazing (see Figure~\ref{fig:SIAT}). The authors  developed a novel Fusion-discriminator which can integrate the frequency information as additional priors and constraints into the dehazing network.

\subsection{neptuneai}

\begin{figure}
\centering
\includegraphics[width=1\linewidth]{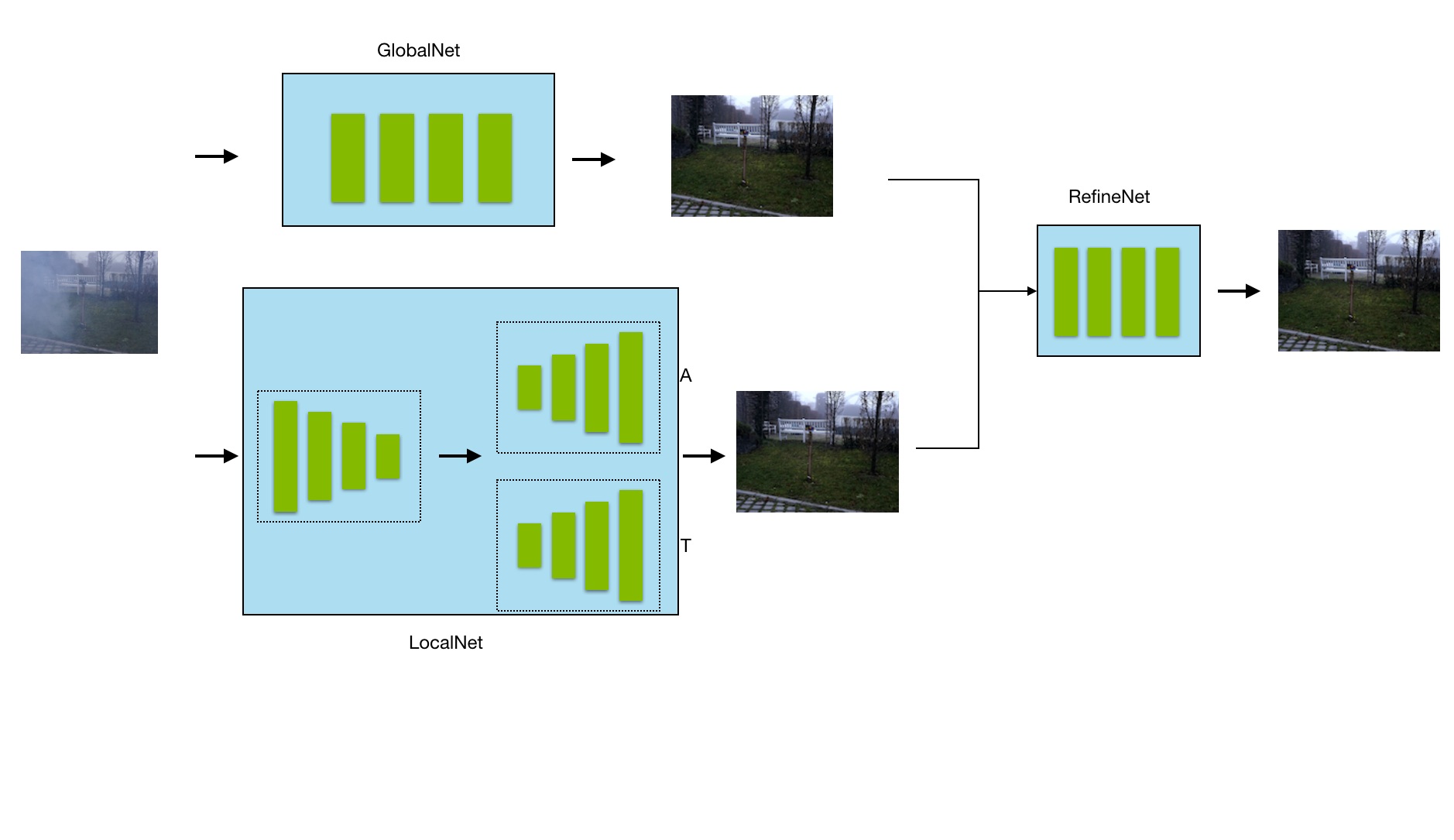}
\caption{The architecture of the proposed model of the \emph{neptuneai} team}
\label{fig:nai}
\end{figure}

The \emph{neptuneai} team used a GlobalNet and LocalNet to generate dehazing results separately. Then the
outputs of these two networks are refined by a RefineNet to get the final clear
image. The GlobalNet is inspired by the AODNet. The LocalNet has a encoder and
two decoders, one for the transmission map and one for the ambient light.
When training the network, a loss for each output of these three
networks was computed. Then, the weighted sum the losses was computed to get the final loss. The loss weights
are set to [0.1, 0.1, 1.0].
The architecture of the proposed model is shown in Figure~\ref{fig:nai}.

\subsection{Neuro-avengers}

\begin{figure}
\centering
\includegraphics[width=1\linewidth]{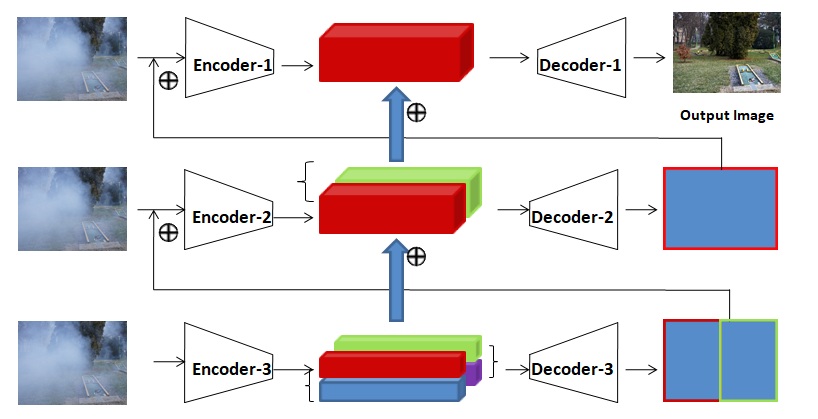}
\caption{The architecture of the proposed model of the Neuro-avengers team}
\label{fig:n-a}
\end{figure}

The proposed  network\cite{Das_fast_deep_2020}
has 3 hierarchies (see Figure~\ref{fig:n-a}). Number of patches used are 1, 2 and 4, respectively, from
top to bottom. In each hierarchy, there is an encoder-decoder pair,
that works on individual patches separately.In all levels, encoder input is
the hazy frame. Decoder output of lower level is added to Encoder input
to the upper level. In addition to this, there are residual connections
between consecutive levels. Main goal of our model is to aggregate features
multiple image patches from different spatial section of the image for better
performance. 
The number of parameters of the proposed encoder-decoder architecture is decreased
due to the residual links in the proposed model, fact that induces a fast dehazing inference.

\subsection{NITREXZ}

The proposed model contains a standard generative adversarial network. The authors used a similar structure of generator introduced in~\cite{qu2019enhanced}.

\subsection{AISAIL}

The core of the proposed solution is built upon the DeBlurGAN, a GAN implementation that is targeting motion blur reduction~\cite{Kupyn_2017}. One key innovation in this technique is the introduction of Correntropy~\cite{Correntropy_2007} based loss function. This loss function was initially introduced to mitigate the VGG related artifacts in deblurGAN. However, this loss function has shown to be effective against non-Gaussian noise. In the proposed technique, the heterogeneous haze is modeled as a type of non-Gaussian noise. This network is hence referred to as DeBlurGAN-C.
The solution includes two steps. In the pre-processing step, initial haze reduction filter was applied to the raw images (both training images and test images). This filter is based on dark channel dehazing technique. However, the airlight estimation in Ancuti~\etal~\cite{Ancuti_ICIP_Dehazing_2018} was adopted to account for the non-uniform airlight estimation. For transmission light estimation, the technique in~\cite{Park_2014} was adopted.
The pre-processed image was then first used to train the DeBlurGAN-C network. The input training data consists of patches of 256$\times$256. These patches were generated by both splitting the original images into the patch size as well as splitting downscaled input images into the patch size. Additionally, scaling factors of 1, 2 and 4 were employed.
For testing, the input images are also first passing through the pre-processing and then processed by the trained DeblurGAN-C model to produce the restored images.

\begin{figure}
\centering
\includegraphics[width=1\linewidth]{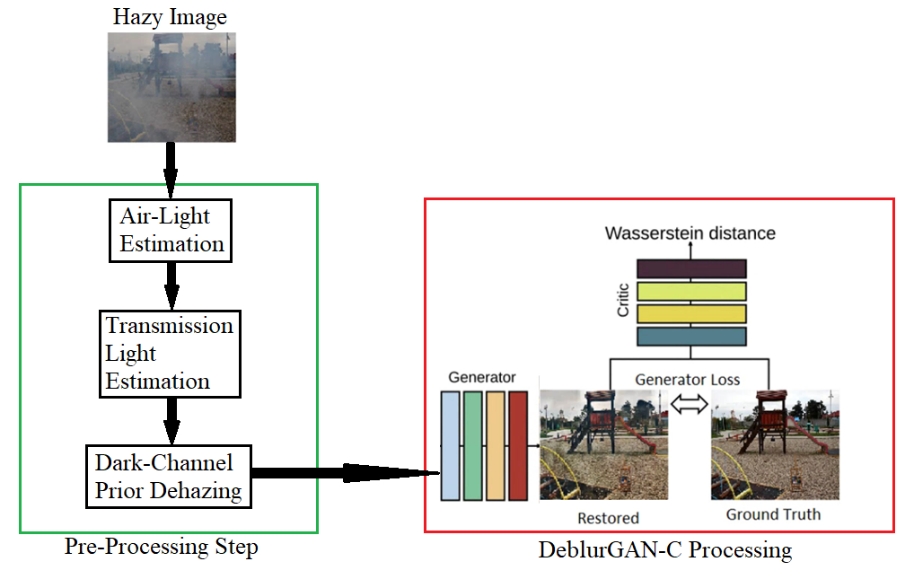}
\caption{The architecture of the proposed model of the AISAIL team}
\label{fig:AISAIL}
\end{figure}

\subsection{ICAIS$\_$dehaze}

\begin{figure}[htbp]
	\centerline{\includegraphics[width=1\linewidth]{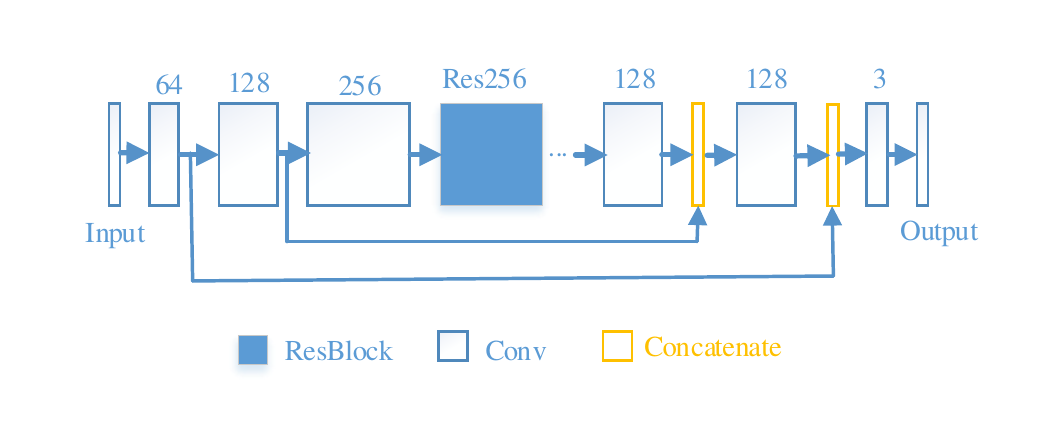}}
	\caption{Overall architecture for the encoder-decoder architecture used in the method of ICAIS$\_$dehaze team. }
	\label{fig:overall}
\end{figure}

This approach uses a  generative adversarial network with the similar structure as CycleGAN. An encoder-decoder architecture with skip connections is introduced in the generator (see Figure~\ref{fig:overall}). Multiple residual blocks are used in both  the encoder and the decoder. The output of the discriminator is downsampled to three scales before calculating the discriminator loss to reconstruct the multi-scale features. GAN loss, perceptual loss and L1 loss are used during the training process. The paired image similarity is ensured by the losses  on both sides, \ie, the loss for hazy transformed to clean and clean to hazy.

\subsection{RETINA}

This approach,  named  spatio-temporal retinex-inspired by an averaging of stochastic samples (STRASS) , is based on the spatio-temporal envelope retinex-inspired with a stochastic sampling framework (STRESS)~\cite{Kolas_2011} and also from the random spray retinex (RSR)~\cite{Provezi_2007}.  In this work, the authors used the idea of the relation developed in~\cite{Dravo_2015} replacing the envelope structure of the samples used in ~\cite{Kolas_2011} by an average of these samples. Due to the local
properties of the algorithm, this modified computation in the framework also impacted regions of the image far from the camera.

\subsection{hazefreeworld}

\begin{figure}[!htb]
		\centering
		\includegraphics[width=1\linewidth]{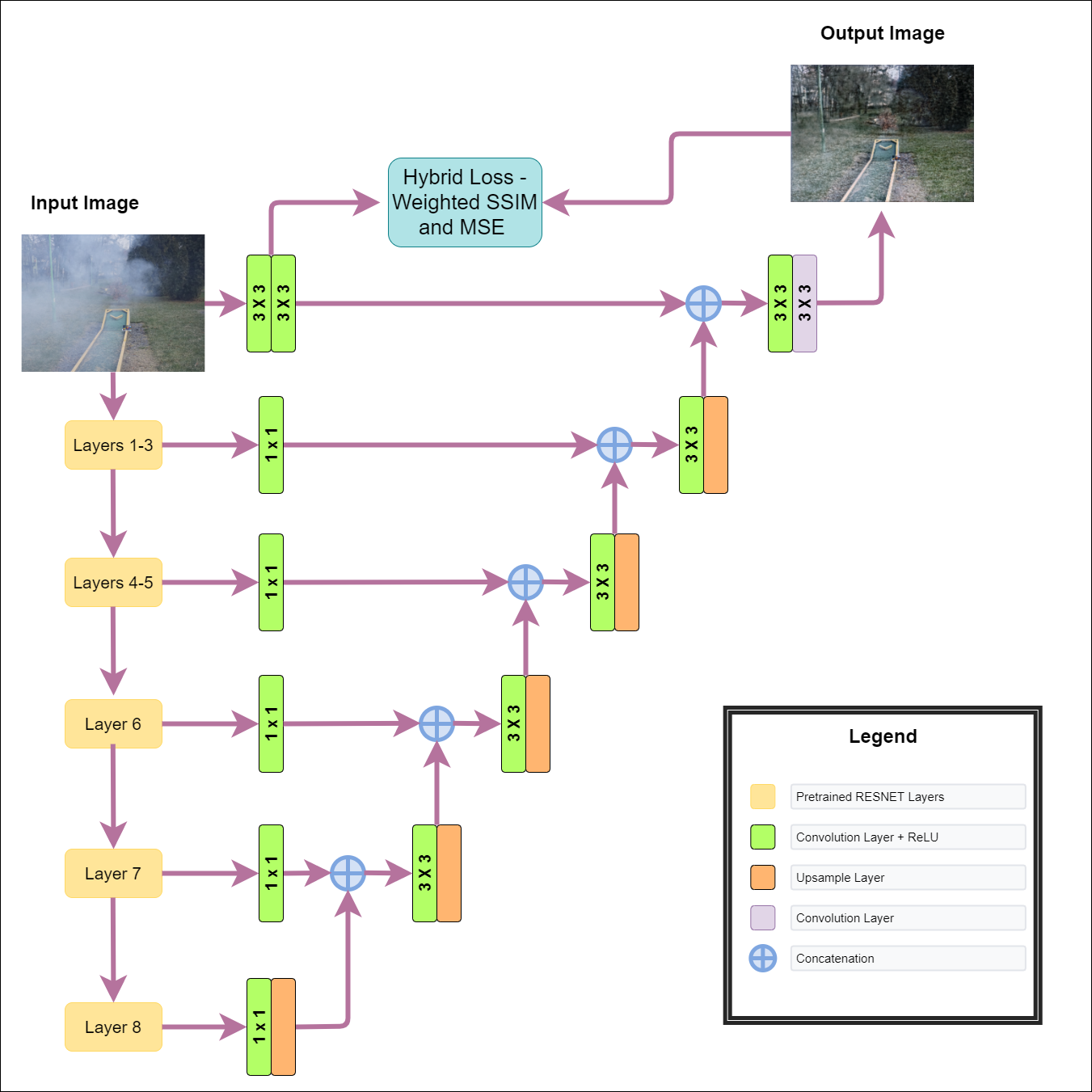}\\
		\caption{The architecture of the method proposed by \textbf{hazefreeworld} team. }
		\label{fig:fig_hf}
\end{figure}

This method utilizes a convolutional neural network architecture based on the skeleton of a U-Net. The proposed network uses the first 8 layers from a pretrained RESNET-18 network (see Figure~\ref{fig:fig_hf}) for efficient encoding. It has been trained on both the NTIRE20 dataset (NH-HAZE dataset~\cite{ancuti2020NH-HAZE}) as well as the following external datasets -- I-HAZE~\cite{Ancuti_IHAZE_2018}, O-HAZE~\cite{Ancuti_OHAZE_2018}, HazeRD and D-HAZY~\cite{D_Hazy_2016}. The custom loss function used is a weighted hybrid loss combining SSIM metric with MSE loss using a weighting factor that reflects their relative magnitude and effect on image quality. The relative weighting of the SSIM Loss to MSE was 0.9999 to 0.0001 based on their relative magnitude and effect on image quality. The skip connections from U-Net ensure that there is no loss in context with respect to the input. The optimizer used for training is Adam with an initial learning rate of 1e-3 and a weight decay of 1e-6. When used on a GPU platform, the model processes images in 0.9486 seconds.

\section*{Acknowledgments}

We thank the NTIRE 2020 sponsors: HUAWEI Technologies Co. Ltd., OPPO Mobile Corp., Ltd., Voyage81, MediaTek Inc., DisneyResearch$\mid$Studios, and ETH Zurich (Computer Vision Lab). Part of this work has been supported by 2020 European Union Research and Innovation Horizon 2020 under the grant agreement Marie Sklodowska-Curie No 712949 (TECNIOspring PLUS), as well as the Agency for the Competitiveness of the Company of the Generalitat de Catalunya - ACCIO: TECSPR17-1-0054.

\appendix
\section{Teams and affiliations}
\label{sec:appendix}
\subsection*{NTIRE 2020 team}
\noindent\textit{\textbf{Title: }} NTIRE 2020 Challenge on Nonhomogeneous Dehazing\\
\noindent\textit{\textbf{Members: }}\textit{Codruta O. Ancuti$^1$ (codruta.ancuti@gmail.com)}, Cosmin Ancuti$^{1,2}$, Florin-Alexandru Vasluianu$^3$, Radu Timofte$^3$\\
\noindent\textit{\textbf{Affiliations: }}\\
$^1$ University Politehnica of Timisoara, Romania\\
$^2$ University of Girona, Spain\\
$^3$ Computer Vision Lab, ETH Zurich, Switzerland\\

\subsection*{ECNU-Trident}

\noindent\textit{\textbf{Title:} Trident Dehazing Network} \\
\noindent\textit{\textbf{Members:}} \textit{Jing Liu (splinter02@163.com)}, Haiyan Wu, Yuan Xie, Yanyun Qu, Lizhuang Ma\\
\noindent\textit{\textbf{Affiliations: }}\\
East China Normal University, Multimedia and Computer Vision Lab\\

\subsection*{ECNU-KT}

\noindent\textit{\textbf{Title:} NonHomogeneous Dehazing via Knowledge Transfer} \\
\noindent\textit{\textbf{Members:}} \textit{Haiyan Wu (704289013@qq.com)}, Jing Liu, Yuan Xie, Yanyun Qu, Lizhuang Ma\\
\noindent\textit{\textbf{Affiliations: }}\\
East China Normal University(ECNU), Multimedia and Computer Vision Lab\\

\subsection*{dehaze\_sneaker}

\noindent\textit{\textbf{Title:} Atmospheric model guided and Haze-Aware Representation Distillation Module for Single Image Dehazing} \\
\noindent\textit{\textbf{Members:}} \textit{Ziling Huang (huangziling@gapp.nthu.edu.tw)},  Qili Deng\\
\noindent\textit{\textbf{Affiliations: }}\\
National TsingHua University/ByteDance AI Lab\\

\subsection*{Spider}

\noindent\textit{\textbf{Title:} Haze-Aware Representation Distillation for Single Image  Dehazing} \\
\noindent\textit{\textbf{Members:}} \textit{Qili Deng (dengqili@bytedance.com), Ziling Huang}\\
\noindent\textit{\textbf{Affiliations: }}\\
ByteDance AI Lab/National Tsing Hua University\\

\subsection*{NTU Dehazing}

\noindent\textit{\textbf{Title:} UNet-based Model for Single Image Dehazing} \\
\noindent\textit{\textbf{Members:}} \textit{Ju-Chin Chao (R07942078@ntu.edu.tw)}, 
Tsung-Shan Yang, Peng-Wen Chen, Po-Min Hsu, Tzu-Yi Liao, Chung-En Sun, Pei-Yuan Wu\\
\noindent\textit{\textbf{Affiliations: }}\\
Graduate Institute of Communication Engineering, National Taiwan University, Taiwan\\

\subsection*{VICLAB-DoNET}

\noindent\textit{\textbf{Title:} Adaptive Attention based U-Net for Coarse-to-Fine Framework} \\
\noindent\textit{\textbf{Members:}} \textit{Jeonghyeok Do (ehwjdgur0913@kaist.ac.kr)}, Jongmin Park, Munchurl Kim\\
\noindent\textit{\textbf{Affiliations: }}\\
Korea Advanced Institute of Science and Technology (KAIST), Daejoeon, Republic of Korea\\

\subsection*{iPAL-NonLocal}

\noindent\textit{\textbf{Title:} NonLocal Channel Attention for NonHomogeneous Dehazing} \\
\noindent\textit{\textbf{Members:}} \textit{Kareem Metwaly (kareem@psu.edu)}, Xuelu Li, Tiantong Guo and Vishal Monga\\
\noindent\textit{\textbf{Affiliations: }}\\
The Pennsylvania State University\\School of Electrical Engineering and Computer Science\\The Information Processing and Algorithms Laboratory (iPAL) \\

\subsection*{Team JJ}

\noindent\textit{\textbf{Title:} DenseNet-based UNet with multi-scale strategy} \\
\noindent\textit{\textbf{Members:}} \textit{Jongmin Park (pjm36671673@kaist.ac.kr)}, Jeonghyeok Do, Munchurl Kim\\
\noindent\textit{\textbf{Affiliations: }}\\
Korea Advanced Institute of Science and Technology (KAIST), Daejoeon, Republic of Korea\\

\subsection*{iPAL-EDN}

\noindent\textit{\textbf{Title:} Dense Ensemble Dehazing Network} \\
\noindent\textit{\textbf{Members:}} \textit{Mingzhao Yu (ethanyu@psu.edu)}, Venkateswararao Cherukuri, Tiantong Guo, Vishal Monga\\
\noindent\textit{\textbf{Affiliations: }}\\
The Pennsylvania State University\\School of Electrical Engineering and Computer Science\\The Information Processing and Algorithms Laboratory (iPAL) \\

\subsection*{NTUEE\_LINLAB}

\noindent\textit{\textbf{Title:} Multi-scale Encoder with Bicycle-GAN using Difference} \\
\noindent\textit{\textbf{Members:}} \textit{Shiue-Yuan Chuang  (sychuang0909@ntu.edu.tw), Tsung-Nan Lin, David Lee, Jerome Chang, Zhan-Han Wang}\\
\noindent\textit{\textbf{Affiliations: }}\\
National Taiwan University\\

\subsection*{NTUST-merg}

\noindent\textit{\textbf{Title:} Dense-DehazeNet} \\
\noindent\textit{\textbf{Members:}} \textit{Yu-Bang Chang  (aes851122@gmail.com), Chang-Hong Lin}\\
\noindent\textit{\textbf{Affiliations: }}\\
National Taiwan University of Science and Technology, Taiwan\\

\subsection*{SIAT}

\noindent\textit{\textbf{Title:} Generative Adversarial Networks with Fusion-discriminator for Single Image Dehazing} \\
\noindent\textit{\textbf{Members:}} \textit{Yu Dong (yudong@siat.ac.cn), Hongyu Zhou}\\
\noindent\textit{\textbf{Affiliations: }}\\
Shenzhen Institutes of Advanced Technology of the Chinese Academy of Science, China\\

\subsection*{neptuneai}

\noindent\textit{\textbf{Title:} Global and local fusion network for image dehazing} \\
\noindent\textit{\textbf{Members:}} \textit{Xiangzhen Kong (neptune.team.ai@gmail.com)}\\
\noindent\textit{\textbf{Affiliations: }}\\\\

\subsection*{Neuro-avengers}

\noindent\textit{\textbf{Title:} Deep Multi-patch Hierarchical Network for Image Dehazing} \\
\noindent\textit{\textbf{Members:}} \textit{Sourya Dipta Das (dipta.juetce@gmail.com)}, Saikat Dutta\\
\noindent\textit{\textbf{Affiliations: }}\\
Jadavpur University and IIT Madras\\

\subsection*{NITREXZ}

\noindent\textit{\textbf{Title:} Haze Model Based Generative Adversarial
Network For Image Dehazing} \\
\noindent\textit{\textbf{Members:}} \textit{Xuan Zhao (zxstud@163.com)}\\
\noindent\textit{\textbf{Affiliations: }}\\
Nanjing University of Aeronautics and Astronautics, Nanjing, China\\

\subsection*{AISAIL}

\noindent\textit{\textbf{Title:} Single image dehazing using GAN and Correntropy loss function} \\
\noindent\textit{\textbf{Members:}} \textit{Bing Ouyang (bouyang@fau.edu)}, Dennis Estrada\\
\noindent\textit{\textbf{Affiliations: }}\\
Harbor Branch Oceanographic Institute/Florida Atlantic University\\

\subsection*{ICAIS$\_$dehaze}

\noindent\textit{\textbf{Title:} A Generative Adversarial Network  for NonHomogeneous Dehazing} \\
\noindent\textit{\textbf{Members:}} \textit{Meiqi Wang (mqwang@smail.nju.edu.cn)},  Tianqi Su, Siyi Chen\\
\noindent\textit{\textbf{Affiliations: }}\\
School of Electronic Science and Engineering, Nanjing University, China\\

\subsection*{RETINA}

\noindent\textit{\textbf{Title:} Spatio-Temporal Retinex-Inspired by an
Averaging of Stochastic Samples (STRASS)} \\
\noindent\textit{\textbf{Members:}} \textit{Bangyong Sun (sunbangyong@xaut.edu.cn)},  Vincent Whannou de Dravo,  Zhe Yu\\
\noindent\textit{\textbf{Affiliations: }}\\
School of Printing, Packaging and Digital Media,
Xi'an University of Technology, Shaanxi, 710048 China\\

\subsection*{hazefreeworld}

\noindent\textit{\textbf{Title:} Deep Non-homogeneous Dehazing with Hybrid Weighted Loss} \\
\noindent\textit{\textbf{Members:}} \textit{Pratik Narang (pratik.narang@pilani.bits-pilani.ac.in)}, Aryan Mehra, Navaneeth Raghunath, Murari
Mandal\\
\noindent\textit{\textbf{Affiliations:} BITS Pilani and MNIT Jaipur}\\
\\

{\small
\bibliographystyle{ieee_fullname}
\bibliography{egbib}

\begin{thebibliography}{10}\itemsep=-1pt

\bibitem{abdelhamed2020ntire}
Abdelrahman Abdelhamed, Mahmoud Afifi, Radu Timofte, Michael Brown, et~al.
\newblock {NTIRE 2020} challenge on real image denoising: Dataset, methods and
  results.
\newblock In {\em The IEEE Conference on Computer Vision and Pattern
  Recognition (CVPR) Workshops}, June 2020.

\bibitem{abdelhamed2019ntire}
Abdelrahman Abdelhamed, Radu Timofte, and Michael~S Brown.
\newblock Ntire 2019 challenge on real image denoising: Methods and results.
\newblock In {\em Proceedings of the IEEE Conference on Computer Vision and
  Pattern Recognition Workshops}, pages 0--0, 2019.

\bibitem{Ancuti_TIP_2013}
C.O. Ancuti and C. Ancuti.
\newblock Single image dehazing by multi-scale fusion.
\newblock {\em IEEE Transactions on Image Processing}, 22(8):3271--3282, 2013.

\bibitem{Ancuti_NTIRE_2018}
C. Ancuti, C.O. Ancuti, R. Timofte, L.~Van Gool, and L.~Zhang et al.
\newblock {NTIRE} 2018 challenge on image dehazing: Methods and results.
\newblock {\em IEEE CVPR, {NTIRE} Workshop}, 2018.

\bibitem{Ancuti_DENSE_HAZE_2019}
C. Ancuti, C.~O. Ancuti, M. Sbert, and R. Timofte.
\newblock {DENSE-HAZE}: A benchmark for image dehazing with dense-haze and
  haze-free images.
\newblock {\em IEEE ICIP}, 2019.

\bibitem{Ancuti_IHAZE_2018}
C. Ancuti, C.~O. Ancuti, R. Timofte, and C. {De Vleeschouwer}.
\newblock {I-HAZE}: a dehazing benchmark with real hazy and haze-free indoor
  images.
\newblock {\em International Conference on Advanced Concepts for Intelligent
  Vision Systems}, 2018.

\bibitem{D_Hazy_2016}
C. Ancuti, C.~O. Ancuti, and Christophe~De Vleeschouwer.
\newblock {D-Hazy}: A dataset to evaluate quantitatively dehazing algorithms.
\newblock {\em IEEE ICIP}, 2016.

\bibitem{Ancuti_ICIP_Dehazing_2018}
C.~O. Ancuti, C. Ancuti, and C. {De Vleeschouwer}.
\newblock Effective local airlight estimation for image dehazing.
\newblock In {\em IEEE ICIP}, 2018.

\bibitem{Ancuti_NT_TIP_2020}
C.~O. Ancuti, C. Ancuti, C. {De Vleeschouwer}, and A.~C. Bovick.
\newblock Day and night-time dehazing by local airlight estimation.
\newblock In {\em IEEE Transactions on Image Processing}, 2020.

\bibitem{Ancuti_OHAZE_2018}
C.~O. Ancuti, C. Ancuti, C. {De Vleeschouwer}, and R. Timofte.
\newblock {O-HAZE}: a dehazing benchmark with real hazy and haze-free outdoor
  images.
\newblock {\em IEEE CVPR, {NTIRE} Workshop}, 2018.

\bibitem{ancuti2020NH-HAZE}
Codruta~O. Ancuti, Cosmin Ancuti, and Radu Timofte.
\newblock {NH-HAZE}: An image dehazing benchmark with nonhomogeneous hazy and
  haze-free images.
\newblock In {\em The IEEE Conference on Computer Vision and Pattern
  Recognition (CVPR) Workshops}, 2020.

\bibitem{ancuti2020ntire}
Codruta~O. Ancuti, Cosmin Ancuti, Florin-Alexandru Vasluianu, Radu Timofte,
  et~al.
\newblock {NTIRE 2020} challenge on nonhomogeneous dehazing.
\newblock In {\em The IEEE Conference on Computer Vision and Pattern
  Recognition (CVPR) Workshops}, June 2020.

\bibitem{Ancuti_NTIRE_2019}
C.~O. Ancuti, C.Ancuti, R. Timofte, L.~Van Gool, and L.~Zhang et al.
\newblock {NTIRE} 2019 challenge on image dehazing: Methods and results.
\newblock {\em IEEE CVPR, {NTIRE} Workshop}, 2019.

\bibitem{arad2020ntire}
Boaz Arad, Radu Timofte, Yi-Tun Lin, Graham Finlayson, Ohad Ben-Shahar, et~al.
\newblock {NTIRE 2020} challenge on spectral reconstruction from an rgb image.
\newblock In {\em The IEEE Conference on Computer Vision and Pattern
  Recognition (CVPR) Workshops}, June 2020.

\bibitem{Blau_2018_ECCV_Workshops}
Yochai Blau, Roey Mechrez, Radu Timofte, Tomer Michaeli, and Lihi Zelnik-Manor.
\newblock The 2018 pirm challenge on perceptual image super-resolution.
\newblock In {\em The European Conference on Computer Vision (ECCV) Workshops},
  September 2018.

\bibitem{Dehazenet_2016}
B. Cai, X. Xu, K. Jia, C. Qing, and D. Tao.
\newblock Dehazenet: An end-to-end system for single image haze removal.
\newblock {\em IEEE Transactions on Image Processing}, 2016.

\bibitem{cai2019ntire}
Jianrui Cai, Shuhang Gu, Radu Timofte, and Lei Zhang.
\newblock Ntire 2019 challenge on real image super-resolution: Methods and
  results.
\newblock In {\em Proceedings of the IEEE Conference on Computer Vision and
  Pattern Recognition Workshops}, pages 0--0, 2019.

\bibitem{Das_fast_deep_2020}
Sourya~Dipta Das and Saikat Dutta.
\newblock Fast deep multi-patch hierarchical network for nonhomogeneous image
  dehazing.
\newblock In {\em The IEEE Conference on Computer Vision and Pattern
  Recognition (CVPR) Workshops}, June 2020.

\bibitem{Dravo_2015}
V.~Whannou~De Dravo and J.~Y. Hardeberg.
\newblock Stress for dehazing.
\newblock In {\em In Colour and Visual Computing Symposium}, 2015.

\bibitem{Fattal_Dehazing}
Raanan Fattal.
\newblock Single image dehazing.
\newblock {\em SIGGRAPH}, 2008.

\bibitem{fuoli2020ntire}
Dario Fuoli, Zhiwu Huang, Martin Danelljan, Radu Timofte, et~al.
\newblock {NTIRE 2020} challenge on video quality mapping: Methods and results.
\newblock In {\em The IEEE Conference on Computer Vision and Pattern
  Recognition (CVPR) Workshops}, June 2020.

\bibitem{gao2019res2net}
Shanghua Gao, Ming-Ming Cheng, Kai Zhao, Xin-Yu Zhang, Ming-Hsuan Yang, and
  Philip~HS Torr.
\newblock Res2net: A new multi-scale backbone architecture.
\newblock {\em IEEE transactions on pattern analysis and machine intelligence},
  2019.

\bibitem{gu2019aim}
Shuhang Gu, Martin Danelljan, Radu Timofte, et~al.
\newblock Aim 2019 challenge on image extreme super-resolution: Methods and
  results.
\newblock In {\em International Conference on Computer Vision Workshop
  (ICCVW)}, pages 3556--3564. IEEE, 2019.

\bibitem{gu2019brief}
Shuhang Gu and Radu Timofte.
\newblock A brief review of image denoising algorithms and beyond.
\newblock In {\em Inpainting and Denoising Challenges}, pages 1--21. Springer,
  2019.

\bibitem{Dehaze_He_CVPR_2009}
K. He, J. Sun, and X. Tang.
\newblock Single image haze removal using dark channel prior.
\newblock {\em In IEEE CVPR}, 2009.

\bibitem{he2016deepresidual}
Kaiming He, Xiangyu Zhang, Shaoqing Ren, and Jian Sun.
\newblock Deep residual learning for image recognition.
\newblock In {\em Proceedings of the IEEE conference on computer vision and
  pattern recognition}, pages 770--778, 2016.

\bibitem{huang_2017_densely}
G. {Huang}, Z. {Liu}, L. v.~d. {Maaten}, and K.~Q. {Weinberger}.
\newblock Densely connected convolutional networks.
\newblock In {\em 2017 IEEE Conference on Computer Vision and Pattern
  Recognition (CVPR)}, pages 2261--2269, July 2017.

\bibitem{DenseNet}
Gao Huang, Zhuang Liu, Laurens van~der Maaten, and Kilian~Q Weinberger.
\newblock Densely connected convolutional networks.
\newblock In {\em Proceedings of the IEEE Conference on Computer Vision and
  Pattern Recognition}, pages 4700--4708, 2017.

\bibitem{ignatov2019ntire}
Andrey Ignatov and Radu Timofte.
\newblock Ntire 2019 challenge on image enhancement: Methods and results.
\newblock In {\em Proceedings of the IEEE Conference on Computer Vision and
  Pattern Recognition Workshops}, pages 0--0, 2019.

\bibitem{Ignatov_2018_ECCV_Workshops}
Andrey Ignatov, Radu Timofte, et~al.
\newblock Pirm challenge on perceptual image enhancement on smartphones:
  Report.
\newblock In {\em The European Conference on Computer Vision (ECCV) Workshops},
  September 2018.

\bibitem{isola2017image}
Phillip Isola, Jun-Yan Zhu, Tinghui Zhou, and Alexei~A Efros.
\newblock Image-to-image translation with conditional adversarial networks.
\newblock In {\em Proceedings of the IEEE conference on computer vision and
  pattern recognition}, pages 1125--1134, 2017.

\bibitem{Kolas_2011}
O. Kolas, I. Farup, and A. Rizzi.
\newblock Spatio-temporal retinex-inspired envelope with stochastic sampling: a
  framework for spatial color al- gorithms.
\newblock In {\em Journal of Imaging Science and Technology}, 2011.

\bibitem{Koschmieder_1924}
H. Koschmieder.
\newblock Theorie der horizontalen sichtweite.
\newblock In {\em Beitrage zur Physik der freien Atmosphare}, 1924.

\bibitem{Kratz_and_Nishino_2009}
L. Kratz and K. Nishino.
\newblock Factorizing scene albedo and depth from a single foggy image.
\newblock {\em ICCV}, 2009.

\bibitem{Kupyn_2017}
O. Kupyn, V. Budzan, M. Mykhailych, D. Mishkin, and J. Matas.
\newblock Deblurgan: Blind motion deblurring using conditional adversarial
  networks.
\newblock {\em arXiv:1711.07064}, 2017.

\bibitem{liu2020trident}
Jing Liu, Haiyan Wu, Yuan Xie, Yanyun Qu, and Lizhuang Ma.
\newblock Trident dehazing network.
\newblock In {\em The IEEE Conference on Computer Vision and Pattern
  Recognition (CVPR) Workshops}, June 2020.

\bibitem{Liu_dehazing_2018}
Q. Liu, X. Gao, L. He, and W. Lu.
\newblock Single image dehazing with depth-aware non-local total variation
  regularization.
\newblock {\em IEEE Trans. Image Proc.}, 2018.

\bibitem{Liu_2019}
R. Liu, L. Ma, Y. Wang, and L. Zhang.
\newblock Learning converged propagations with deep prior ensemble for image
  enhancement.
\newblock {\em IEEE Trans. Image Proc.}, 2019.

\bibitem{Correntropy_2007}
W. Liu, P.~P. Pokharel, and J.~C. Principe.
\newblock Correntropy: properties and applications in non-gaussian signal
  processing.
\newblock {\em IEEE Transactions on Signal Processing}, 2007.

\bibitem{lugmayr2020ntire}
Andreas Lugmayr, Martin Danelljan, Radu Timofte, et~al.
\newblock {NTIRE 2020} challenge on real-world image super-resolution: Methods
  and results.
\newblock In {\em The IEEE Conference on Computer Vision and Pattern
  Recognition (CVPR) Workshops}, June 2020.

\bibitem{lugmayr2019aim}
Andreas Lugmayr, Martin Danelljan, Radu Timofte, Manuel Fritsche, Shuhang Gu,
  et~al.
\newblock {AIM} 2019 challenge on real-world image super-resolution: Methods
  and results.
\newblock In {\em International Conference on Computer Vision Workshop
  (ICCVW)}. IEEE, 2019.

\bibitem{mccartney1976optics}
Earl~J McCartney.
\newblock Optics of the atmosphere: scattering by molecules and particles.
\newblock {\em New York, John Wiley and Sons, Inc., 1976. 421 p.}, 1976.

\bibitem{Metwaly_NonLocal_2020}
Kareem Metwaly, Xuelu Li, Tiantong Guo, and Vishal Monga.
\newblock Nonlocal channel attention for nonhomogeneous image dehazing.
\newblock In {\em The IEEE Conference on Computer Vision and Pattern
  Recognition (CVPR) Workshops}, June 2020.

\bibitem{nah2020ntire}
Seungjun Nah, Sanghyun Son, Radu Timofte, Kyoung~Mu Lee, et~al.
\newblock {NTIRE} 2020 challenge on image and video deblurring.
\newblock In {\em The IEEE Conference on Computer Vision and Pattern
  Recognition (CVPR) Workshops}, June 2020.

\bibitem{Park_2014}
H. Park, D. Park, D.~K. Han, and H. Ko.
\newblock Single image haze removal using novel estimation of atmospheric light
  and transmission.
\newblock In {\em IEEE ICIP}, 2014.

\bibitem{Provezi_2007}
E. Provenzi, M. Fierro, A. Rizzi, L. Carli, D. Gadia, and D. Marini.
\newblock Random spray retinex: A new retinex implementation to investi- gate
  the local properties of the model.
\newblock In {\em IEEE Transactions on Image Processing}, 2007.

\bibitem{qin2019ffa}
Xu Qin, Zhilin Wang, Yuanchao Bai, Xiaodong Xie, and Huizhu Jia.
\newblock Ffa-net: Feature fusion attention network for single image dehazing.
\newblock {\em arXiv preprint arXiv:1911.07559}, 2019.

\bibitem{qu2019enhanced}
Yanyun Qu, Yizi Chen, Jingying Huang, and Yuan Xie.
\newblock Enhanced pix2pix dehazing network.
\newblock In {\em Proceedings of the IEEE Conference on Computer Vision and
  Pattern Recognition}, pages 8160--8168, 2019.

\bibitem{Ren_2016}
W. Ren, S. Liu, H. Zhang, J. Pan, X. Cao, and M.-H. Yang.
\newblock Single image dehazing via multi-scale convolutional neural networks.
\newblock {\em Proc. European Conf. Computer Vision}, 2016.

\bibitem{ronneberger2015unet}
Olaf Ronneberger, Philipp Fischer, and Thomas Brox.
\newblock U-net: Convolutional networks for biomedical image segmentation.
\newblock In {\em International Conference on Medical image computing and
  computer-assisted intervention}, pages 234--241. Springer, 2015.

\bibitem{shi2016real}
Wenzhe Shi, Jose Caballero, Ferenc Husz{\'a}r, Johannes Totz, Andrew~P Aitken,
  Rob Bishop, Daniel Rueckert, and Zehan Wang.
\newblock Real-time single image and video super-resolution using an efficient
  sub-pixel convolutional neural network.
\newblock In {\em IEEE Conference on Computer Vision and Pattern Recognition},
  pages 1874--1883, 2016.

\bibitem{szegedy2017inception}
Christian Szegedy, Sergey Ioffe, Vincent Vanhoucke, and Alexander~A Alemi.
\newblock Inception-v4, inception-resnet and the impact of residual connections
  on learning.
\newblock In {\em Thirty-First AAAI Conference on Artificial Intelligence},
  2017.

\bibitem{Tan_Dehazing}
Robby~T. Tan.
\newblock Visibility in bad weather from a single image.
\newblock {\em In IEEE Conference on Computer Vision and Pattern Recognition},
  2008.

\bibitem{Tarel_ICCV_2009}
J.-P. Tarel and N. Hautiere.
\newblock Fast visibility restoration from a single color or gray level image.
\newblock {\em In IEEE ICCV}, 2009.

\bibitem{Tarel_2012}
J.-P. Tarel, N. Hautière, L. Caraffa, A. Cord, H. Halmaoui, and D. Gruyer.
\newblock Vision enhancement in homogeneous and heterogeneous fog.
\newblock {\em IEEE Intelligent Transportation Systems Magazine}, 2012.

\bibitem{Timofte_2017_CVPR_Workshops}
Radu Timofte, Eirikur Agustsson, Luc Van~Gool, Ming-Hsuan Yang, Lei Zhang,
  et~al.
\newblock Ntire 2017 challenge on single image super-resolution: Methods and
  results.
\newblock In {\em The IEEE Conference on Computer Vision and Pattern
  Recognition (CVPR) Workshops}, July 2017.

\bibitem{Timofte_2018_CVPR_Workshops}
Radu Timofte, Shuhang Gu, Jiqing Wu, and Luc Van~Gool.
\newblock Ntire 2018 challenge on single image super-resolution: Methods and
  results.
\newblock In {\em The IEEE Conference on Computer Vision and Pattern
  Recognition (CVPR) Workshops}, June 2018.

\bibitem{Timofte_2016_CVPR}
Radu Timofte, Rasmus Rothe, and Luc Van~Gool.
\newblock Seven ways to improve example-based single image super resolution.
\newblock In {\em The IEEE Conference on Computer Vision and Pattern
  Recognition (CVPR)}, June 2016.

\bibitem{ulyanov2016instance}
Dmitry Ulyanov, Andrea Vedaldi, and Victor Lempitsky.
\newblock Instance normalization: The missing ingredient for fast stylization.
\newblock {\em arXiv preprint arXiv:1607.08022}, 2016.

\bibitem{Wang_2019}
A. Wang, W. Wang, J. Liu, and N. Gu.
\newblock {AIPNet}: Image-to-image single image dehazing with atmospheric
  illumination prior.
\newblock {\em IEEE Trans. Image Proc.}, 2019.

\bibitem{wang2018esrgan}
Xintao Wang, Ke Yu, Shixiang Wu, Jinjin Gu, Yihao Liu, Chao Dong, Yu Qiao, and
  Chen Change~Loy.
\newblock Esrgan: Enhanced super-resolution generative adversarial networks.
\newblock In {\em Proceedings of the European Conference on Computer Vision
  (ECCV)}, pages 0--0, 2018.

\bibitem{wu2020knowledge}
Haiyan Wu, Jing Liu, Yuan Xie, Yanyun Qu, and Lizhuang Ma.
\newblock Knowledge transfer dehazing network for nonhomogeneous dehazing.
\newblock In {\em The IEEE Conference on Computer Vision and Pattern
  Recognition (CVPR) Workshops}, June 2020.

\bibitem{yu2018wide}
Jiahui Yu, Yuchen Fan, Jianchao Yang, Ning Xu, Xinchao Wang, and Thomas~S
  Huang.
\newblock Wide activation for efficient and accurate image super-resolution.
\newblock {\em arXiv preprint arXiv:1808.08718}, 2018.

\bibitem{Yu_ensemble_2020}
Mingzhao Yu, Venkateswararao Cherukuri, Tiantong Guo, and Vishal Monga.
\newblock Ensemble dehazing networks for non-homogeneous haze.
\newblock In {\em The IEEE Conference on Computer Vision and Pattern
  Recognition (CVPR) Workshops}, June 2020.

\bibitem{yuan2020demoireing}
Shanxin Yuan, Radu Timofte, Ales Leonardis, Gregory Slabaugh, et~al.
\newblock {NTIRE 2020} challenge on image demoireing: Methods and results.
\newblock In {\em The IEEE Conference on Computer Vision and Pattern
  Recognition (CVPR) Workshops}, June 2020.

\bibitem{Zhang_dehazing_2018}
H. Zhang, V. Sindagi, and V.~M. Patel.
\newblock Multi-scale single image dehazing using perceptual pyramid deep
  network.
\newblock {\em IEEE CVPR}, 2018.

\bibitem{zhang2020ntire}
Kai Zhang, Shuhang Gu, Radu Timofte, et~al.
\newblock {NTIRE 2020} challenge on perceptual extreme super-resolution:
  Methods and results.
\newblock In {\em The IEEE Conference on Computer Vision and Pattern
  Recognition (CVPR) Workshops}, June 2020.

\bibitem{zhang2018perceptual}
Richard Zhang, Phillip Isola, Alexei~A Efros, Eli Shechtman, and Oliver Wang.
\newblock The unreasonable effectiveness of deep features as a perceptual
  metric.
\newblock In {\em CVPR}, 2018.

\bibitem{zhang2018unreasonable}
Richard Zhang, Phillip Isola, Alexei~A Efros, Eli Shechtman, and Oliver Wang.
\newblock The unreasonable effectiveness of deep features as a perceptual
  metric.
\newblock In {\em IEEE Conference on Computer Vision and Pattern Recognition},
  pages 586--595, 2018.

\bibitem{zhu2019deformable}
Xizhou Zhu, Han Hu, Stephen Lin, and Jifeng Dai.
\newblock Deformable convnets v2: More deformable, better results.
\newblock In {\em Proceedings of the IEEE Conference on Computer Vision and
  Pattern Recognition}, pages 9308--9316, 2019.

\end{thebibliography}
}

\end{document}